\documentclass{article} % For LaTeX2e
\usepackage{iclr2026_conference,times}

\usepackage{amsmath,amsfonts,bm}

\def\eqref#1{equation~\ref{#1}}
\def\1{\bm{1}}

\DeclareMathAlphabet{\mathsfit}{\encodingdefault}{\sfdefault}{m}{sl}
\SetMathAlphabet{\mathsfit}{bold}{\encodingdefault}{\sfdefault}{bx}{n}

\DeclareMathOperator*{\argmin}{arg\,min}

\usepackage{hyperref}
\usepackage{url}

\usepackage[dvipsnames]{xcolor}

\usepackage{multirow} % table cell combination
\usepackage{amsmath} % gather
\usepackage{bbm} % indicator function
\usepackage{siunitx} % angle
\usepackage{kotex} % korean
\usepackage{array,multirow,graphicx}
\usepackage{float}
\usepackage{amsmath}
\usepackage{mathtools}
\usepackage{bbm}
\usepackage{hhline}
\usepackage{boldline}
\usepackage{tabularx}
\usepackage{float}
\usepackage{comment}
\usepackage{color}
\usepackage{amssymb}
\usepackage{booktabs}
\usepackage{multicol}
\usepackage{microtype}      % microtypography
\usepackage{xcolor}         % colors
\usepackage{algorithm}
\usepackage{algpseudocode}
\usepackage{pifont}% http://ctan.org/pkg/pifont
\usepackage{soul} %\st
\usepackage{subcaption}
\usepackage{wrapfig}
\usepackage{lipsum}
\usepackage{booktabs}
\usepackage{xcolor,colortbl}
\usepackage{nicematrix}
\usepackage{kotex}
\usepackage{dsfont}
\usepackage{caption}
\usepackage[accsupp]{axessibility}
\usepackage{amsmath}
\title{From Vicious to Virtuous Cycles: \\ Synergistic Representation Learning for \\ Unsupervised Video Object-Centric Learning}

\author{Hyun Seok Seong\thanks{Equal Contribution} \;\;\; WonJun Moon\footnotemark[1] \;\;\; Jae-Pil Heo\thanks{Corresponding Author} \\
Sungkyunkwan University\\
\texttt{\{gustjrdl95, wjun0830, jaepilheo\}@skku.edu}\\
}
\iclrfinalcopy % Uncomment for camera-ready version, but NOT for submission.
\begin{document}

\maketitle

\begin{abstract}
Unsupervised object-centric learning models, particularly slot-based architectures, have shown great promise in decomposing complex scenes. 
However, their reliance on reconstruction-based training creates a fundamental conflict between the sharp, high-frequency attention maps of the encoder and the spatially consistent but blurry reconstruction maps of the decoder. 
We identify that this discrepancy gives rise to a vicious cycle\textcolor{blue}{:} the noisy feature map from the encoder forces the decoder to average over possibilities and produce even blurrier outputs, while the gradient computed from blurry reconstruction maps lacks high-frequency details necessary to supervise encoder features.
To break this cycle, we introduce Synergistic Representation Learning~(SRL) that establishes a virtuous cycle where the encoder and decoder mutually refine one another. 
SRL leverages the encoder's sharpness to deblur the semantic boundary within the decoder output, while exploiting the decoder's spatial consistency to denoise the encoder's features.
This mutual refinement process is stabilized by a warm-up phase with a slot regularization objective that initially allocates distinct entities per slot.
By bridging the representational gap between the encoder and decoder, SRL achieves state-of-the-art results on video object-centric learning benchmarks.
Codes are available at \href{https://github.com/hynnsk/SRL}{github.com/hynnsk/SRL}.
% \blfootnote{
% $^\dagger$ Equal contribution} 
% \blfootnote{
% $^\ast$ Corresponding author
% }
\end{abstract}
\section{Introduction}
Object-centric representation learning aims to decompose complex scenes into a set of disentangled object representations, a critical capability for robust video understanding~\citep{slotvlm}. 
Among prevailing approaches, slot-based models~\citep{slotattention, slotcontrast, videosaur, savi, elsayed2022savi++} have demonstrated significant promise in learning to group pixels into meaningful object-level slots in an unsupervised manner. 
These models typically operate by encoding a video into a feature map, which is then parsed by an attention mechanism into a fixed number of latent slots. 
The quality of these slots is subsequently evaluated by a decoder that attempts to reconstruct the original input from them, using a reconstruction loss like Mean Squared Error~(MSE) as the primary training signal. 
This reconstruction-based supervision is vital, as it circumvents the need for manual annotations and provides a workable objective in a purely unsupervised setting where direct supervision on feature grouping is noisy and difficult to formulate.

However, we identify a fundamental discrepancy inherent in this widely adopted training paradigm. 
The learning process relies on two distinct spatial maps that are unfortunately misaligned in their characteristics: (1)~the attention maps generated by the slot attention, and (2)~the decoded output maps produced by the reconstruction decoder. 
The attention maps, derived from pixel-wise feature similarities, are inherently sharp and granular, but also susceptible to high-frequency noise. 
In contrast, the decoded output maps, typically generated by passing the flattened slots through an MLP decoder, tend to be blurry and spatially smooth. 
This blurring effect is an artifact of the autoencoder's architecture~(e.g., Slot Attention) and the smoothing nature of the MSE loss~\citep{mseblur1, mseblur2}, which leads to perceptual compression~\citep{rombach2022high}.

\begin{figure*}[t]
\centering
\vspace{-0.3cm}
\includegraphics[width=0.91\columnwidth]{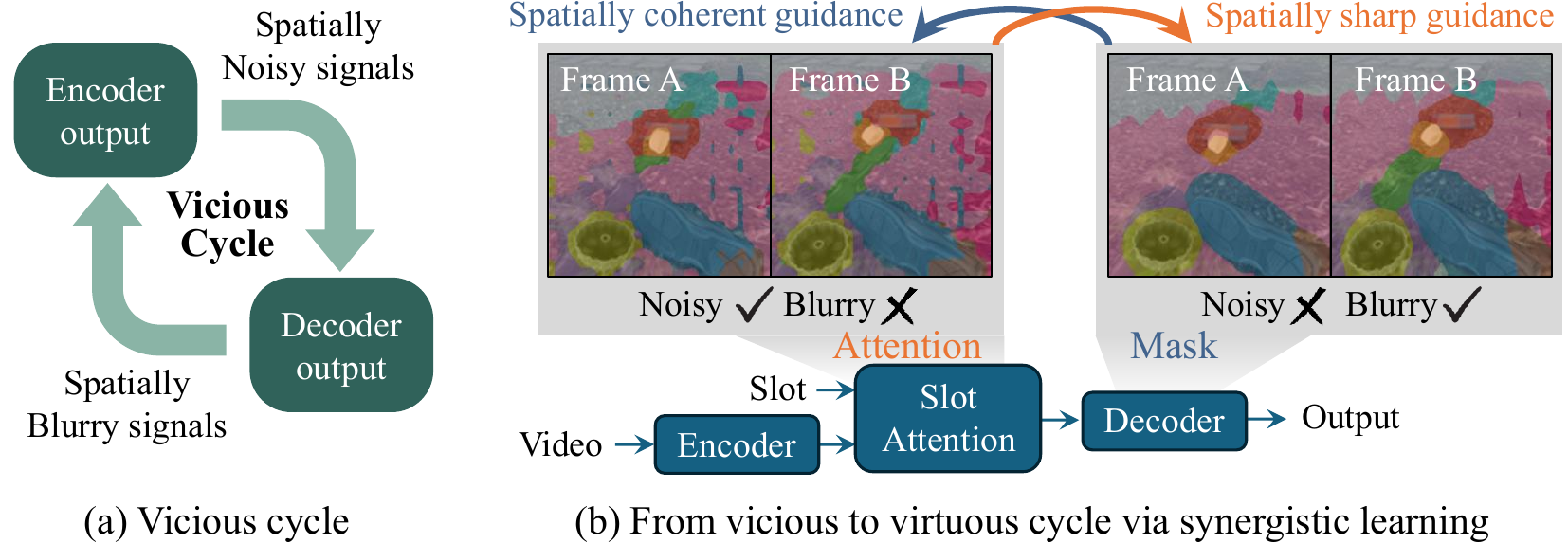} 
\vspace{-0.2cm}
\caption{
    (a) Vicious cycle in video object-centric learning.
    Noisy inputs from the encoder render the decoder's reconstruction task ill-posed, reinforcing its tendency to produce blurry, low-frequency outputs. In turn, the corrupted gradient from these blurry outputs lacks the high-frequency detail required to refine the encoder's sharp but noisy features.
    (b) Virtuous cycle of synergistic representation learning.
    Our framework transforms this conflict into collaboration.
    We leverage the encoder's sharp attention maps to deblur the decoder output while denoising the encoder features with the decoder's spatially coherent masks.
}
\label{fig1.motivation}
\vspace{-0.3cm}
\end{figure*}

This discrepancy incurs a vicious feedback loop that fundamentally constrains the learning process, as shown in Fig.~\ref{fig1.motivation}.
On one hand, the encoder, while leveraging sharp DINO-v2 features~\citep{Dinov2}, produces noisy groupings by incorrectly associating spatially distant patches~\citep{denoisingVIT}. 
When the decoder receives these noisy slot representations, its reconstruction task becomes ill-posed. 
To minimize the MSE penalty under this uncertainty, the decoder’s safest strategy is to average over the possibilities, which further reinforces its own tendency to produce blurry outputs by a biased optimization toward recovering low-frequency content.
On the other hand, such a decoder provides a corrupted, low-frequency learning signal to the rest of the model. 
The gradients flowing back to the encoder lack the precise high-frequency details necessary to supervise the learning of sharp encoder features.

To break this vicious cycle and establish a virtuous one, we introduce Synergistic Representation Learning, a novel framework where the two spatial maps synergistically refine one another through purpose-built objectives. 
First, we tackle the decoder's blurriness by leveraging the encoder's sharp, albeit noisy, attention map as a guide. 
We introduce a ternary contrastive objective for deblurring that strategically partitions patches into three tiers: the anchor itself, other patches grouped with the anchor by the encoder's sharp attention map, and all other patches.
A ranking loss then enforces this objective, compelling the decoder to resolve ambiguities at the object boundary where its blurry grouping conflicts with the encoder's sharp prior.
On the other hand, we leverage the decoder’s more spatially coherent representation to provide a training signal to denoise the encoder’s noisy representations.
Specifically, we exploit another ternary contrastive objective to use the decoder’s consistent masks to enforce spatial consistency within encoded feature maps, pushing spuriously grouped, distant patches apart in the feature space. 
This entire refinement process is built upon a robust warm-up phase that employs a slot regularization loss. 
This prevents the initial slot collapse by identifying and resetting redundant slots, ensuring a meaningful foundation for the subsequent decoder deblurring and encoder denoising processes.
Through this co-evolving optimization process, our method successfully bridges the gap between the noisy encoder and the blurry decoder, resulting in significantly sharper object segmentation and more robust representations. 
% We demonstrate that our approach achieves state-of-the-art performance on video object-centric learning benchmarks.

\section{Related Work}

\subsection{Object-Centric Representation Learning}
The goal of object-centric learning is to decompose complex scenes into a set of discrete object representations without explicit supervision~\citep{smm, OCL4, OCL1, ocl5, ocl3}. 
A significant breakthrough in this area is Slot Attention~\citep{slotattention}, which employs an iterative, competitive attention mechanism to bind slots to different objects in an image. 
Each slot, initialized randomly, refines its representation over several iterations by competing for evidence from the input features, effectively performing a soft, differentiable version of clustering.

This paradigm was successfully extended to the temporal domain for video processing. 
Earlier works like SAVi~\citep{savi, elsayed2022savi++} and STEVE~\citep{steve} maintain temporal consistency by propagating slot representations from one frame to the next, enabling robust unsupervised object tracking and decomposition in dynamic scenes.
Subsequently, Videosaur~\citep{videosaur} proposed a self-supervised task to predict patch motion, and more recently, SlotContrast~\citep{slotcontrast} introduced slot-level contrastive learning between slots of successive frames to enhance temporal consistency.

Other popular streams include reducing the redundancy in slot representations and self-distillation.
To reduce the redundancy between slots, SOLV~\citep{aydemir2023self} merges slots via a non-differentiable agglomerative clustering procedure, while MetaSlot~\citep{metaslot} addresses redundancy by using a codebook to prune duplicated slots. 
While these approaches are effective, they rely on explicit redundancy detection, which evolves relatively slowly during training.
Similarly, we employ a slot regularization objective to mitigate redundancy. 
However, a key difference is that our regularization is aggressively applied only during the initial training iterations.
% This allows the slots exhibit well-separated representations that serve as a robust foundation for the subsequent representation learning process
This allows the encoder-decoder architecture to achieve stable representations early in training, establishing a robust foundation for the subsequent learning process.
% This allows the encoder-decoder architecture to focus exclusively on slot representation learning after a brief warm-up period, achieving distinct and well-separated representations that serve as a robust foundation for the subsequent representation learning process.

The other stream of research focuses on self-distillation~\citep{spot, dias}. 
For instance, SPOT~\citep{spot} distills decoder signals into encoder attention, while DIAS~\citep{dias} transfers later-iteration attention to earlier steps. 
While effective in certain settings, these methods directly imitate teacher attention signals without explicitly addressing the noise inherent in the teacher's knowledge. 
In contrast, our synergistic representation learning aims to leverage only the complementary strengths of encoder and decoder representations. 
We mitigate the impact of noisy signals by stratifying mutual signals into intermediate levels, rather than enforcing strong positive or negative constraints.

\subsection{Contrastive Representation Learning}
% \citep{infonce}
% \citep{}
% To address this misalignment, we leverage contrastive representation learning~\citep{}. 
% The goal of contrastive learning is to learn an embedding space where semantically similar~(positive) samples are pulled closer together, while dissimilar~(negative) samples are pushed apart. 
% This is often achieved by minimizing the InfoNCE loss~\citep{infonce}.
% Contrastive Learning의 supervised version(cite supcon)은 multiple positive가 효과적임을 보여준다. 따라서 unsupervised task에서도 semantically similar sample을 찾아서 positive set으로 사용하면 feature clustering에 효과적이다 (cite: 이미지 단위에서 top-k contrastive learning 논문, HP, CAUSE). 따라서 우리는 unsupervised video learning에서 semantic alignment를 frame 내에서, 그리고 frame 사이에서 시키기 위해 positive pressure를 induce하는 contrastive learning을 수행한다.
% 또한 contrastive learning에서는 hard negative를 gather하는게 중요하다(cite). 우리는 Object-Centric Learning framework에서 핵심적인 두 가지 mask인 (attn, decoder mask)의 각각의 약점을 보완할 수 있는 hard negative 를 gather 하여 학습하는 방법을 고안하였다.

Contrastive representation learning~\citep{simclr, moco, byol, swav} has emerged as a powerful paradigm for learning discriminative embeddings by encouraging semantically similar samples to be mapped closer together while pushing apart dissimilar ones. 
This objective is typically instantiated through the InfoNCE loss~\citep{infonce}, which enforces such pairwise alignment in the embedding space. 
Building upon this foundation, supervised extensions~\citep{supcon, kang2020exploring} have demonstrated the effectiveness of leveraging multiple positives per anchor, showing that clustering semantically consistent samples enhances representation quality.

Extending this idea to the unsupervised setting, several works have explored strategies for mining semantically similar samples, such as selecting top-$K$ nearest neighbors as positives~\citep{dwibedi2021little, HP}. 
These approaches demonstrate that expanding the set of positives beyond simple augmentations leads to more robust feature clustering. 
Parallel to positive mining, another line of work has emphasized the importance of hard negative selection, showing that the quality of hard negatives is crucial for effective contrastive learning~\citep{robinson2020hard, kalantidis2020hard}.

Inspired by these insights, we introduce a contrastive framework that leverages the complementary conflict between the encoder's sharp but noisy features and the decoder's coherent but blurry masks. 
By defining a ternary structure of patch relationships, we weaponize this discrepancy: the encoder's sharpness provides a deblurring signal for the decoder, while the decoder's coherence provides a denoising signal for the encoder. 
This cross-source hard negative mining compels each module to overcome its weaknesses, resulting in a synergistic refinement of object-centric representations.

% Inspired by these insights, we employ contrastive learning to promote semantic alignment in video object-centric learning, gathering multiple positives to induce semantic consistency both within frames and across temporal spans. 
% To further enhance representation quality, we leverage the complementary weakness of the two spatial maps. Specifically, encoder features, while sharp, are often noisy, and decoder masks, while coherent, tend to be blurry. 
% We therefore draw hard negatives for the decoder from the encoder’s sharp attention maps, and conversely, obtain hard negatives for the encoder from the decoder’s spatially consistent masks. 
% This cross-source hard negative mining compels each module to overcome its inherent limitations, leading to a synergistic refinement of object-centric representations.

% within the object-centric learning framework, we identify disagreement regions between two complementary spatial masks, attention maps and decoder masks, as a rich source of informative hard negatives. 
% By exploiting these disagreement regions, our method gathers hard negatives that highlight the weaknesses of each mask, enabling them to mutually compensate and ultimately improve object-centric representations.
\section{Method}

% 아래 related work에서 자세히 설명하는걸로.
% \subsection{Preliminary}
% \subsubsection{Video Slot Attention}
% Object-centric representation learning aims to decompose a scene into a set of object-specific slots. 
% Slot Attention~\citep{slotattention} achieves this through an iterative, competitive attention mechanism. 
% Given a set of $K$ randomly initialized slots and input features from an image encoder, the model performs multiple rounds of attention where slots are updated via a Gated Recurrent Unit~(GRU)~\citep{gru}. 
% In each round, slots compete to explain different parts of the input, effectively clustering the input features into object-centric representations.

% In the video domain, this mechanism is extended to enforce temporal consistency. 
% Models like SAVi~\citep{savi} employ a recurrent framework where the slots from frame $t−1$ are used to initialize the slots for frame $t$, allowing the model to track objects through time. 
% The entire model is typically trained end-to-end by decoding the final slot representations and minimizing a pixel-wise reconstruction loss, such as Mean Squared Error~(MSE), against the input video frames.

\begin{figure*}[t]
\centering
\vspace{-10pt}
\includegraphics[width=0.95\columnwidth]{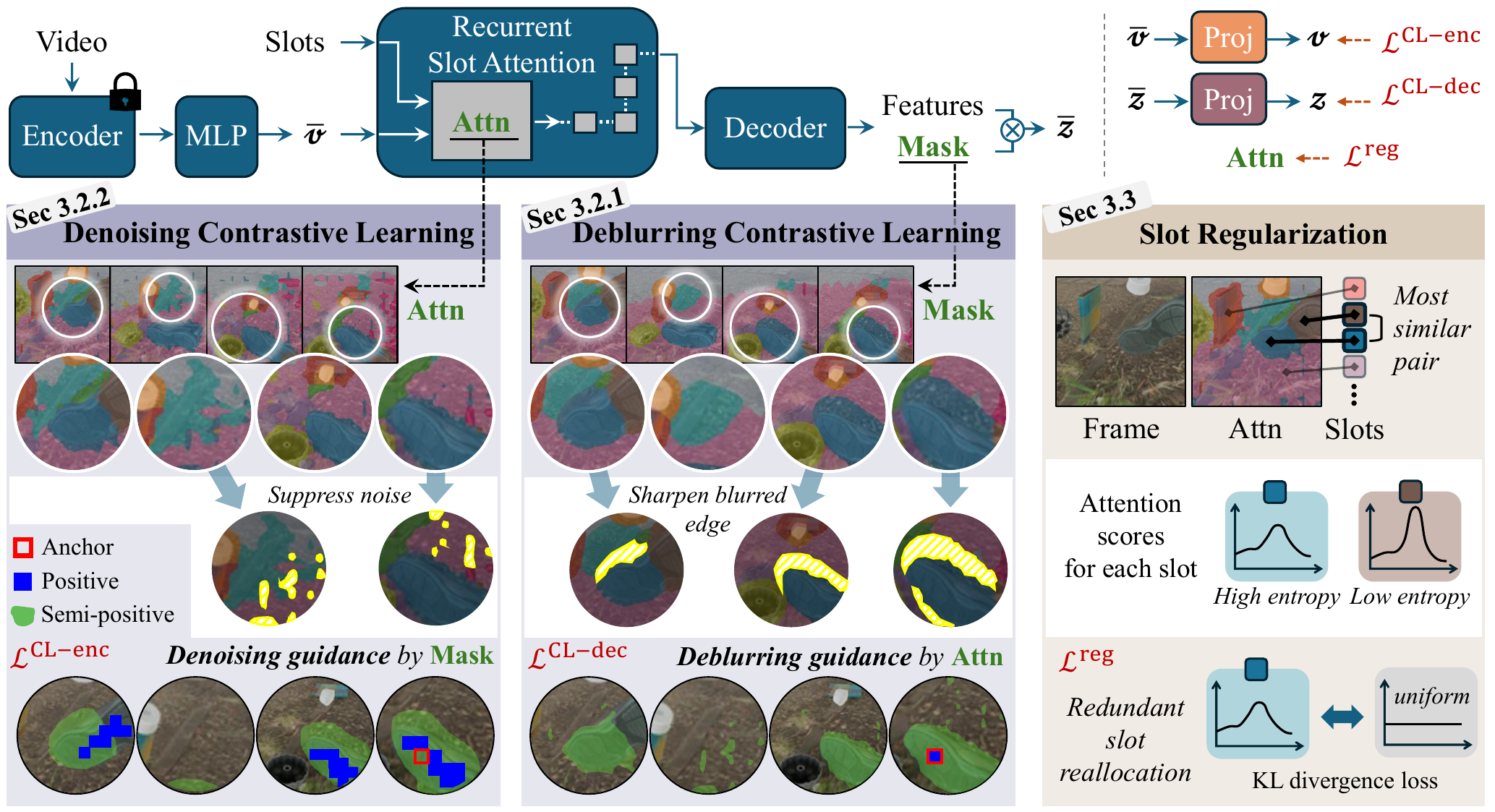} 
\vspace{-0.1cm}
\caption{Overview of Synergistic Representation learning.
    The typical pipeline~(top) suffers from a conflict between the encoder's sharp but noisy features~($\bar{\bm{v}}$) and the decoder's spatially coherent but blurry features~($\bar{\bm{z}}$).
    Our framework breaks this cycle by forcing the two modules to synergistically refine one another: (1) Deblurring path: Encoder's sharp attention map is used to refine the blurry decoded features and (2) Denoising path: Decoder's coherent masks provide a robust signal to denoise the encoder's noisy features.
    Finally, slot regularization during warm-up establishes a solid foundation for this process by ensuring diverse slot specialization.
}
\vspace{-0.3cm}
\label{fig1.main}
\end{figure*}

We address a vicious cycle in video object-centric learning caused by a representational conflict between the slot attention maps and decoded output masks after reconstruction. % reconstruction maps
This conflict recursively inhibits optimization, preventing the learning of clean object representations.
% To break this, we introduce Synergistic Representation Learning~(SRL). 
% As presented in Fig.~\ref{fig1.main}, SRL establishes a virtuous cycle where the two maps mutually refine each other via contrastive losses. 
% This process is stabilized by a warm-up phase that promotes robust slot specialization, laying a strong foundation for the mutual refinement.
To break this, we introduce Synergistic Representation Learning~(SRL), which combines mutual refinement process via contrastive losses with a slot-regularization warm-up phase.
As presented in Fig.~\ref{fig1.main}, SRL initially uses slot regularization to promote robust and non-redundant slot specialization, and then establishes a virtuous cycle in which the slot attention maps and decoded masks iteratively refine each other through contrastive learning.
% We address a fundamental limitation in video object-centric learning: a representational discrepancy between the two crucial spatial maps~(i.e., slot attention map and decoded reconstruction map).
% This discrepancy creates a vicious cycle of learning, where the weaknesses of each module recursively inhibit the optimization of the other, fundamentally constraining the model's ability to learn crisp and clean object representations.
% To break this cycle, we introduce Synergistic Representation Learning~(SRL).
% The overview of SRL is presented in Fig.~\ref{main}.
% In SRL, two distinct spatial maps synergistically improve one another via purpose-built contrastive objectives.
% This virtuous cycle is stabilized by an initial warm-u/p phase that promotes semantic slot specialization, ensuring a robust foundation for the refinement process.

% To break this cycle, we introduce Synergistic Representation Learning~(SRL), where two distinct spatial maps synergistically improve one another. 
% Our framework instantiates this concept via two purpose-built contrastive learning objectives. 
% First, the decoder leverages the encoder's high-frequency sharpness to deblur its own object boundaries. 
% Second, the encoder leverages the decoder's resulting spatial consistency to denoise its own feature representations. 
% This virtuous cycle is stabilized by an initial warm-up phase that promotes semantic slot specialization, ensuring a robust foundation for the refinement process.

\subsection{Vicious Cycle of Encoder-Decoder Discrepancy}
The standard paradigm for training video slot attention models relies on a reconstruction objective~\citep{videosaur, slotattention}, creating an inherent conflict between the encoder's grouping mechanism and the decoder's learning signal. 
This conflict establishes a vicious cycle that hinders optimization.

\textbf{Noisy Encoder Reinforces the Decoder's Blurriness.}
The encoder's features, which are passed to the Slot Attention, are derived from projecting the DINO-v2~\citep{Dinov2, ibot} features, which are designed to be discriminative at a fine-grained level. 
This results in representations that are sharp but also susceptible to noise, such as incorrectly grouping spatially distant patches~\citep{denoisingVIT}.
Subsequently, when the decoder is conditioned on a limited number of noisy latent slots, the reconstruction task becomes ill-posed.
% When the decoder receives these noisy, polluted slot representations, its task of accurate reconstruction becomes ill-posed.
To minimize the MSE loss under this uncertainty, the decoder's safest strategy is to average over the possibilities, further reinforcing its tendency to produce blurry, over-smoothed outputs.
% This is because under MSE, the loss is calculated from the square of the error, meaning that the penalty grows quadratically with the error's magnitude.
% This is because under MSE, the loss is calculated from the square of the error, meaning that an incorrect guess results in a disproportionately larger penalty.

\textbf{Blurry Decoder Corrupts the Encoder's Learning Signal.}
Conversely, the decoder is trained with a pixel-wise objective like MSE, which inherently acts as a low-pass filter. 
It incentivizes the model to predict the conditional expectation of pixel values, resulting in reconstructions that are spatially smooth but suffer from blurry object boundaries and a loss of high-frequency detail. 
The gradient signal that trains the entire model, including the encoder, originates from this blurry output. 
Consequently, the encoder receives a signal that fails to provide the precise guidance needed to learn sharp object boundaries and refine the noisy patch representations.
%the encoder 
%Consequently, the encoder receives a corrupted, low-frequency signal that fails to provide the precise guidance needed to learn sharp object boundaries.

\subsection{From Vicious to Virtuous Cycles via Synergistic Representation Learning}
We aim to break this vicious cycle with SRL, equipped with two purpose-built contrastive objectives that force the encoder and decoder to refine their representations mutually.
Our representation learning begins after a warmup stage, allowing the model to leverage two distinct spatial maps: the slot attention map~($\textbf{Attn} \in \mathbb{R}^{S\times T\times N}$) that is sharp yet noisy, and the decoded mask~($\textbf{Mask} \in \mathbb{R}^{S\times T\times N}$) that is spatially consistent yet blurry.
$S, T$, and $N$ denote the number of slots, frames, and spatial patches.
From these probabilistic maps, we derive hard pseudo-semantic labels.
For a given frame $t$ and spatial patch $i$, pseudo-semantic labels are defined as:
% \begin{equation}
%     l^{\textbf{Attn}} = \argmax_{S} \textbf{Attn} \; ; \; l^{\textbf{Mask}} = \argmax_{S} \textbf{Mask}.
% \end{equation}
% \begin{equation}
%     l^{\textbf{Attn}}(t,i) = \argmax_{s\in \{1,...,S\}} \textbf{Attn}(s,t,i) \; ; \; l^{\textbf{Mask}(t,i)} = \argmax_{s\in\{1,...,S \}} \textbf{Mask}(s,t,i).
% \end{equation}
\begin{equation}
l^{\textbf{Attn}}_{t,i} = \underset{s \in \{1, \dots, S\}}{\arg\max} \textbf{Attn}_{s,t,i} \quad ; \quad l^{\textbf{Mask}}_{t,i} = \underset{s \in \{1, \dots, S\}}{\arg\max}  \textbf{Mask}_{s,t,i}.
\end{equation}
The label $l^{\textbf{Attn}}$ represents the index of the slot that gives the highest attention score to a specific patch, while $l^{\textbf{Mask}}$ represents the index of the slot with the highest value in the decoded mask for that patch.

\subsubsection{Deblurring Contrastive Learning: \\
\quad\quad\quad\quad\quad\quad\quad\quad\quad Refining the Decoder via Encoder Sharpness}
To counteract the decoder’s blurry object boundaries, we introduce a contrastive objective, $\mathcal{L}_{\text{CL-dec}}$.
Our key insight is to weaponize the representational discrepancy between the modules, using the encoder's sharp, attention-derived groupings to mine hard negatives that explicitly penalize blurriness in the decoder's representation space.
To align our contrastive objective with the reconstruction loss, we formulate the contrastive task as aligning the decoded features with their corresponding features from the backbone encoder, which also serves as the target guidance for the reconstruction loss.
Following typical conventions~\citep{simclr, supcon}, we project both feature sets into another embedding space, where a contrastive loss is then calculated.
% To keep the objective consistent with the reconstruction loss~\citep{mbo}, we project both the backbone features and the decoded features into a shared embedding space and use the backbone features as the target for the contrastive loss.

To guide the decoder toward producing crisp, high-fidelity object boundaries, we organize ternary sets of patches for each anchor patch that explicitly preserves the reliability ordering of patch relationships.
%For each $i$-th anchor patch, the index set $\mathcal{U}$ of all patches across $T$ frames are partitioned into (1) the positive set $\mathcal{P}_i^{\text{dec}}$, (2) the semi-positive set $\mathcal{Q}_i^{\text{dec}}$, and (3) the negative set $\mathcal{N}_i^{\text{dec}}$:
%\begin{equation}
%\mathcal{U} = \mathcal{P}_i^{\text{dec}} \cup \mathcal{Q}_i^{\text{dec}} \cup \mathcal{N}_i^{\text{dec}}.
%\end{equation}
For each anchor patch, identified by its spatio-temporal index $(t, i)$ where $t$ is the frame and $i$ is the spatial location, we partition the universal set of all patch indices $\mathcal{U}$. 
This universal set $\mathcal{U}$ contains all possible index pairs $(t', j)$ across all frames $t' \in \{1,\dots,T\}$ and all flattened spatial locations $j \in \{1,\dots,N\}$.
The set $\mathcal{U}$ is partitioned into three disjoint subsets relative to the anchor $(t, i)$: (1) the positive set $\mathcal{P}_{t,i}^{\text{dec}}$, (2) the semi-positive set $\mathcal{Q}_{t,i}^{\text{dec}}$, and (3) the negative set $\mathcal{N}_{t,i}^{\text{dec}}$:
\begin{equation}
\mathcal{U} = \mathcal{P}_{t,i}^{\text{dec}} \cup \mathcal{Q}_{t,i}^{\text{dec}} \cup \mathcal{N}_{t,i}^{\text{dec}}.
\end{equation}

First, the positive set is defined as the anchor itself ($\mathcal{P}_{t,i}^{\text{dec}} = \{(t,i)\}$).
This anchors the objective to self-reconstruction with the aim of semantic deblurring.
Yet, na\"ively treating all other patches as negatives may corrupt the semantic boundaries by erroneously pushing away patches belonging to the same object.
Therefore, we further distinguish semantically similar and organize the semi-positive set, using the encoder's sharp, attention-derived labels~($l^{\textbf{Attn}}$) as:
\begin{equation}
\mathcal{Q}_{t,i}^{\text{dec}} = \{\ (t', j) \mid l_{t',j}^{\textbf{Attn}} = l_{t,i}^{\textbf{Attn}} \ \},
\end{equation}
and its purpose is to further guide the decoder to learn the structural prior from the encoder.
Lastly, the negative set $\mathcal{N}_{t,i}^{\text{dec}}$ is defined as the complement, containing all semantically distinct patches.
Crucially, this set also includes patches from blurred object boundaries, which the decoder might incorrectly associate with the anchor's object.
By including these ambiguous boundary patches in the negative set, we create a contrastive pressure that further compels the decoder to learn a more deblurred representation of the object.

To enforce this hierarchical structure, we adopt a ranking contrastive loss~\citep{hoffmann2022ranking} that operates on two levels. 
This preserves the reliability ordering by ensuring the anchor is pulled more strongly to its own ground-truth than to its semi-positives, and to its semi-positives more strongly than to its negatives. 
Formally, our decoder contrastive loss $\mathcal{L}^{\text{CL-dec}}$ is expressed as:
%\begin{align}
%\label{eq_contrastive_loss_dec}
%    \mathcal{L}^{\text{CL-dec}}_{t,i} =&-
%    \frac{1}{|\mathcal{P}_{t,i}^{\text{dec}}|}\sum_{p \in \mathcal{P}_{t,i}^\text{dec}} \text{log} \frac{\exp(\bm{z}_{t,i} \cdot \bm{y}_p / \tau)}{\sum\limits_{n \in \mathcal{Q}_{t,i}^{\text{dec}} \cup \mathcal{N}_{t,i}^{\text{dec}}} \exp(\bm{z}_{t,i} \cdot \bm{y}_n / \tau)} \\
%    &-
%    \frac{1}{|\mathcal{Q}_{t,i}^{\text{dec}}|}\sum_{p \in \mathcal{Q}_{t,i}^\text{dec}} \text{log} \frac{\exp(\bm{z}_{t,i} \cdot \bm{y}_p / \tau)}{\sum\limits_{n \in \mathcal{N}_{t,i}^{\text{dec}}} \exp(\bm{z}_{t,i} \cdot \bm{y}_n / \tau)},
%\end{align}
\begin{equation}
\label{eq_contrastive_loss_dec}
\!\mathcal{L}^{\text{CL-dec}}_{t,i} =\!-
\log \frac{\exp(\bm{z}_{t,i} \cdot \bm{y}_{t,i} / \tau)}{\!\!\sum\limits_{n \in \mathcal{Q}_{t,i}^{\text{dec}} \cup \mathcal{N}_{t,i}^{\text{dec}}} \exp(\bm{z}_{t,i} \cdot \bm{y}_n / \tau)}
-
\frac{1}{|\mathcal{Q}_{t,i}^{\text{dec}}|}\sum_{q \in \mathcal{Q}_{t,i}^\text{dec}} \log \frac{\exp(\bm{z}_{t,i} \cdot \bm{y}_q / \tau)}{\!\!\sum\limits_{n \in \mathcal{N}_{t,i}^{\text{dec}}} \exp(\bm{z}_{t,i} \cdot \bm{y}_n / \tau)},
\end{equation}
where $\cdot$ denotes cosine similarity, $\tau$ is a temperature parameter.
$\bm{z}\in \mathbb{R}^{T\times N \times C}$ and $\bm{y}\in \mathbb{R}^{T\times N \times C}$ are projected vectors of decoder and backbone features, respectively, where $C$ is the channel dimension.
This structured signal directly counteracts the low-pass filtering effect of the MSE loss, guiding the decoder to produce crisp object boundaries.

\subsubsection{Denoising Contrastive Learning: \\ \quad\quad\quad\quad\quad\quad\quad\quad\quad Refining the Encoder via Decoder Coherence}
Conversely, while the encoder's features inherited from a powerful backbone~(e.g., DINO-v2~\citep{Dinov2}) are spatially sharp, they are susceptible to assigning high similarity to spurious, far-flung patches~(i.e., noise). 
To resolve this, we leverage the decoder's spatially coherent masks to denoise the encoded MLP features $\bar{\bm{v}}$. 
We instantiate this with a second ternary contrastive objective, $\mathcal{L}^{\text{CL-enc}}$.

The objective's structure is similar to that of $\mathcal{L}^{\text{CL-dec}}$, partitioning patches into positive, semi-positive, and negative sets. 
However, the sets are defined differently to serve the specific goal of denoising rather than sharpening.
To illustrate, for a given anchor patch $(t,i)$, the positive set $\mathcal{P}_{t,i}^{\text{enc}}$ is defined by leveraging the semantic similarity within the backbone.
% To illustrate, the positive set $\mathcal{P}_{t,i}^{\text{enc}}$ is defined by leveraging the semantic similarity within the backbone.
It comprises the anchor's Top-$K$ nearest neighbors in the DINO-v2 embedding space, sampled from all $T$ frames of the video.
This anchors the representation to the strongest semantic signals provided by the backbone, grouping similar patches. 
The semi-positive set $\mathcal{Q}_{t,i}^{\text{enc}}$ is gathered to enforce spatial coherence.
It is defined using the coarse~(blurred) but contiguous object masks generated by the decoder as follows:
\begin{equation}
\mathcal{Q}_{t,i}^{\text{enc}} = \{\ (t',j) \mid (l_{t',j}^{\textbf{Mask}} = l_{t,i}^{\textbf{Mask}}) \ \},
\end{equation}
where all patches that share the same decoder-derived label~($l^{\textbf{Mask}}$) form this set.
The negative set $\mathcal{N}_{t,i}^{\text{enc}}$ is defined as the complement.

Then, we apply the same ranking loss as in Eq.~\ref{eq_contrastive_loss_dec}, except that the projected decoder $\bm{\hat{y}}$ and backbone features $\bm{y}$ are both replaced by the projected MLP features $\bm{v}$.
Then, the objective is expressed as:
\begin{equation}
\label{eq_contrastive_loss_mlp}
    \begin{aligned}
    \mathcal{L}^{\text{CL-enc}}_{t,i} =&-
    \frac{1}{|\mathcal{P}_{t,i}^\text{enc}|}\sum_{p \in \mathcal{P}_{t,i}^{\text{enc}}} \text{log} \frac{\exp(\bm{v}_{t,i} \cdot \bm{v}_p / \tau)}{\sum\limits_{n \in \mathcal{Q}_{t,i}^{\text{enc}} \cup \mathcal{N}_{t,i}^{\text{enc}}} \exp(\bm{v}_{t,i} \cdot \bm{v}_n / \tau)} \\
    &-
    \frac{1}{|\mathcal{Q}_{t,i}^{\text{enc}}|}\sum_{p \in \mathcal{Q}_{t,i}^{\text{enc}}} \text{log} \frac{\exp(\bm{v}_{t,i} \cdot \bm{v}_p / \tau)}{\sum\limits_{n \in \mathcal{N}_{t,i}^{\text{enc}}} \exp(\bm{v}_{t,i} \cdot \bm{v}_n / \tau)}
    .
    \end{aligned}
\end{equation}
% where/ $\bm{v}$ denotes the projected MLP features.
This formulation uses the larger positive set to ensure features of the same class cluster together, while the decoder-derived semi-positive set tightens this cluster around a spatially coherent instance, effectively exposing the noise patches in the negative set.

\subsection{Slot Regularization for Redundancy Reduction}
% A robust initial assignment of slots to objects is a critical prerequisite for our mutual refinement process. 
% Our contrastive objectives operate at a fine-grained level, calibrating slot representations only after the slots have already captured the coarse semantics of distinct objects. 
% To avoid the degenerate case where multiple slots redundantly attend to the same object~(fragmenting different parts) in pursuit of minimizing reconstruction error, we introduce a slot regularization phase during warm-up.

A reliable initial assignment of slots to objects is a critical prerequisite for our mutual refinement process. 
Our contrastive objectives operate at a fine-grained level and are intended to calibrate slot representations after the slots have already captured the coarse semantics of distinct objects. 
However, in practice, objects are often fragmented into multiple slots when the model aggressively minimizes reconstruction error, leading to several redundant slots covering the same object region. 
In such cases, these spatially overlapping slots may continue to cooperate and further fragment the object instead of consolidating it. 
To prevent this degenerate behavior, we introduce a slot regularization objective.

This regularization identifies and penalizes $M$ most redundant slots, iteratively performing the following steps $M$ times.
First, the model identifies the most similar slot pair, ($\hat{i}$,$\hat{j}$), by finding the pair of indices that maximizes the cosine similarity between their final representations at frame $T$:
\begin{equation}
\label{eq:max-pair}
(\hat{i}, \hat{j}) = \underset{1 \le i < j \le S}{\text{argmax}} \;(\bm{s}_{T,i} \cdot \bm{s}_{T,j}).
\end{equation}
%First, the model identifies the most similar pair of slots ($\hat{i}$,$\hat{j}$), by calculating the cosine similarity between their final representations $\bm{s}_{T,a}$ and $\bm{s}_{T,b}$, at frame $T$:
%\begin{equation}
%\label{eq:max-pair}
%(\hat{i}, \hat{j}) = \argmax_{\,1 \le a < b \le S} (\bm{s}_{T,a} \cdot \bm{s}_{T,b}).
%\end{equation}
For the identified pair, it assesses which slot is less specialized to specific semantics. Specialization is measured by mean KL divergence between attention maps for a given slot $m$, denoted as $\textbf{Attn}_{m}$, and a uniform distribution \textbf{U} across all $T$ frames. 
% For the identified pair, it assesses which slot is less specialized to specific semantics. 
% Specialization is measured by the mean KL divergence between a slot's attention maps $\textbf{Attn}_{1:T}$ and a uniform distribution \textbf{U} across all $T$ frames. 
The slot with the lower score is selected for regularization:
\begin{align}
\label{eq:lower-kl-slot}
m^{\text{low}} & = \argmin_{m \in \{\hat{i}, \hat{j}\}}
\frac{1}{T} \sum_{t=1}^{T} D^{\mathrm{KL}}\left( \textbf{Attn}_{m,t} \middle | \textbf{U} \right).
\end{align}
The index of the chosen slot, $m^{\text{low}}$, is then added to a set of penalized indices, $\mathcal{M}^{pen}$.
This slot is subsequently excluded from consideration in the remaining selection steps.
% The chosen slot $m^{\text{low}}$ is then added to the set of $M$ slots to be penalized /and is excluded from consideration in subsequent selection steps.

After this iterative process populates the set $\mathcal{M}^{\text{pen}}$ with $M$ slot indices, we regularize the corresponding slots.
The attention distribution of each penalized slot is encouraged to align with a uniform distribution $\textbf{U}$ via the following KL divergence loss:
\begin{equation}
\label{eq:kl-uniform-loss}
\mathcal{L}^{\text{reg}}
= \frac{1}{M T} \sum_{m \in \mathcal{M}_{\text{pen}}} \sum_{t=1}^{T}
D^{\mathrm{KL}}\left(\mathbf{Attn}_{m,t} \middle| \mathbf{U}\right),
\end{equation}
where $\mathbf{Attn}_{m,t}$ is the attention map of the specific penalized slot with index $m$ at frame $t$. 
This warm-up regularization encourages redundant, less-specialized slots to abandon their overlap and instead discover unexplained regions of the scene, thereby laying a strong foundation for the subsequent mutual refinement.
See Appendix A.2 for a visual illustration of the warm-up effect.
% After $M$ slots have been selected, each slot is regularized by aligning its attention distribution with a uniform distribution $\textbf{U}$ via a final KL divergence loss:
% \begin{equation}
% \label{eq:kl-uniform-loss}
% \mathcal{L}^{\text{reg}}
% = \frac{1}{M T} \sum_{m=1}^{M} \sum_{t=1}^{T}
% D^{\mathrm{KL}}\left(\tilde{\textbf{Attn}}\middle | \textbf{U}\right),
% \end{equation}
% where $\tilde{\textbf{Attn}}$ represents the attention maps of the selected slots. 
% This warm-up regularization encourages redundant, less-specialized slots to abandon their overlap and instead discover unexplained regions of the scene, thereby laying a strong foundation for the subsequent mutual refinement.

\subsection{Staged Training Framework}
Along with our proposed objectives, we adopt the baseline loss, $\mathcal{L}^{\text{base}}$, from SlotContrast~\citep{slotcontrast}, which is composed of the MSE reconstruction loss and the slot-level contrastive loss.
% Our training objective combines our proposed losses with established baseline objectives. 
% Specifically, we adopt the baseline loss, $\mathcal{L}^{\text{base}}$, from SlotContrast~\citep{slotcontrast}, which is composed of the MSE reconstruction loss and the slot-level contrastive loss.
The baseline objectives are applied throughout the training, while our proposed losses are progressively activated as the model's internal representations become more structured.

Specifically, training proceeds in three stages: (i) Slot specialization~(0-10\%), (ii) Slot stabilization~(10-20\%), and (iii) Contrastive refinement~(20-100\%).
During the slot specialization stage, we introduce the regularization loss $\mathcal{L}^{\text{reg}}$, which encourages early semantic differentiation among slots.  
Then, we train solely with $\mathcal{L}^{\text{base}}$ to consolidate stable slot representations.
Finally, we activate our core contribution, $\mathcal{L}^{\text{CL}}$, as both the encoder and decoder representations are sufficiently meaningful for their discrepancy to serve as a rich, structured learning signal at this stage.
The overall objective is as:
\begin{equation}
    \mathcal{L} = \mathcal{L}^{\text{base}} + \mathcal{L}^{\text{stage}}\;\;;\;\; \mathcal{L}^{\text{stage}} =
\begin{cases}
\lambda^{\text{reg}}\mathcal{L}^{\text{reg}}, & \text{if }\eta < 0.1, \\
0, & \text{if } 0.1 \leq \eta < 0.2, \\
\lambda^{\text{CL}}\mathcal{L}^{\text{CL}}, & \text{if } \eta \geq 0.2, \\
\end{cases}
\end{equation}
where $\lambda^{\text{reg}}$ and $\lambda^{\text{CL}}$ are loss coefficients, $\eta$ is a ratio of training progress, and $\mathcal{L}^{\text{CL}} = \mathcal{L}^{\text{CL-enc}} + \mathcal{L}^{\text{CL-dec}}$.

\section{Experiment}

\begin{table}[t]
\centering
\small
\vspace{-10pt}
\caption{Experimental results.
Results are averaged across 3 runs. $\dagger$ is our reproduced version.
% We report averaged results over 3 runs, with $\dagger$ marking our reproduced version.
}
\label{tab:results}
\vspace{-6pt}
\renewcommand{\arraystretch}{0.9}  % Default value: 1
\setlength{\tabcolsep}{5pt} % Default value: 6pt
\begin{tabular}{l cc cc cc}
\toprule
\multirow{2}{*}{\textbf{Method}} &
\multicolumn{2}{c}{\textbf{MOVi‑C}} &
\multicolumn{2}{c}{\textbf{MOVi‑E}} & \multicolumn{2}{c}{\textbf{YouTube-VIS}} \\
\cmidrule(lr){2-3}\cmidrule(lr){4-5}\cmidrule(lr){6-7}
& FG‑ARI\,$\uparrow$ & mBO\,$\uparrow$
& FG‑ARI\,$\uparrow$ & mBO\,$\uparrow$
& FG‑ARI\,$\uparrow$ & mBO\,$\uparrow$ \\ 
\midrule
SAVi~\citep{savi} & 22.2 & 13.6 & 42.8 & 16.0 & - & - \\
STEVE~\citep{steve} & 36.1 & 26.5 & 50.6 & 26.6 & 15.0 & 19.1 \\
VideoSAUR~\citep{videosaur} & 64.8 & \textbf{38.9} & 73.9 & \textbf{35.6} & 28.9 & 26.3 \\
VideoSAURv2~\citep{slotcontrast} & --   & --   & 77.1 & 34.4 & 31.2 & 29.7 \\
SlotContrast~\citep{slotcontrast} & 69.3 & 32.7 & 82.9 & 29.2 & 38.0 & 33.7 \\ 
SlotContrast$^\dagger$~\citep{slotcontrast} & 70.4 & 31.7 & 80.9 & 28.2 & 36.2 & 32.9 \\
\rowcolor{gray!20}
% hs190 & 73.0 & 36.7 & - & - & - & -  \\
% 1GPU | MLP top8, masks ignore warmup 10000 | slot unifirom last k=5 10000 | dec flearn
% Ours & \textbf{74.7} & 34.8 & 81.4 & 29.8 & \textbf{44.1} & \textbf{38.0}  \\ %
SRL~(Ours) & \textbf{74.3} & 34.5 & \textbf{81.9} & 29.3 & \textbf{42.9} & \textbf{35.6}  \\ %
% SRL~(Ours) & \textbf{74.7} & 34.8 & \textbf{82.2} & 29.2 & \textbf{42.9} & \textbf{35.6}  \\ %
\bottomrule
\end{tabular}
\vspace{-12pt}
\end{table}

\subsection{Experiment Settings}
\textbf{Datasets.}
For evaluation, we employ two synthetic and one real-world dataset. 
The synthetic datasets~\citep{greff2022kubric} consist of numerous moving objects placed against complex backgrounds. 
MOVi-C contains up to 11 objects, whereas MOVi-E extends this to 23 objects and additionally incorporates linear camera motion. 
For real-world dataset, we adopt the YouTube-VIS~(YTVIS) 2021~\citep{vis2021}, which provides a diverse collection of video scenes sourced from YouTube.
%For real-world evaluation, we adopt the YouTube-VIS~(YTVIS) 2021~\citep{vis2021} dataset, which provides a diverse collection of video scenes sourced from YouTube.

\textbf{Evaluation Metrics.}
We evaluate object discovery using Foreground Adjusted Rand Index~(FG-ARI) and mean Best Overlap~(mBO)~\citep{mbo}. FG-ARI measures how well predicted masks align with ground-truth objects, excluding background pixels, and reflects segmentation quality and temporal consistency when computed over full videos. mBO, based on intersection-over-union (IoU), matches each ground-truth mask with the best corresponding prediction and averages the IoU, thereby assessing how accurately masks fit object boundaries. For both metrics, we report video-level scores~(capturing consistency across time) as well as image-level scores~(computed per frame).

% We assess the quality and consistency of discovered object masks using two widely adopted metrics in the object-centric learning literature: Foreground Adjusted Rand Index (FG-ARI) and mean Best Overlap (mBO).

% FG-ARI is an adaptation of the standard Adjusted Rand Index that excludes the background mask. It measures the similarity between predicted object masks and ground-truth masks, focusing on how well objects are segmented and separated. In the video setting, FG-ARI is computed across entire sequences, thereby capturing the temporal consistency of object discovery. We also report Image FG-ARI, which is calculated independently for each frame and then averaged, to isolate per-frame segmentation quality from video-level consistency effects.

% mBO evaluates the alignment of predicted and ground-truth masks based on the intersection-over-union (IoU). For each ground-truth mask, the predicted mask with the highest IoU is selected, and the mean IoU over all matched pairs is reported. Unlike FG-ARI, mBO includes background pixels in the evaluation, offering a more precise measure of how accurately masks fit object boundaries. Similar to FG-ARI, we distinguish between Video mBO (computed over entire sequences, reflecting mask consistency) and Image mBO (computed frame by frame).

\textbf{Implementation Details.}
% MOVI dataset과 YTVIS datset의 input video size와 ViT patch 개수
% slot warmup iter
% feature learning start iter
% number of positive (k)
% number of slot uniform (half라 제외)
Following SlotContrast~\citep{slotcontrast}, we employ DINO-v2~\citep{Dinov2} as our backbone, using DINO-v2-Small/14 for MOVi-C and DINO-v2-Base/14 for MOVi-E and YTVIS. 
We resize input frames to 336$\times$336 for MOVi datasets and 518$\times$518 for YTVIS.
% Following previous work~\citep{slotcontrast}, we employ DINO-v2~\citep{Dinov2} as our backbone to extract the images, resized to 336x336 and 518x518 for MOVi and YTVIS datasets, respectively.
We set the number of positive samples per anchor in denoising contrastive learning~($K$) to $8T$ for MOVi-C, $24T$ for MOVi-E, and $16T$ for YTVIS, respectively.
The number of slots is set to 11, 15, and 7 for each dataset, following SlotContrast~\citep{slotcontrast}.
The number of penalized slots~($M$) was consistently set to half the total slot count~($M=\lfloor S/2 \rfloor$), and loss coefficients $\lambda^{\text{CL}}$ and $\lambda^{\text{reg}}$ are set to 0.1 for all datasets.
% Our training protocol is divided into three stages.
% We begin with a slot specialization warm-up phase~(0-10\% of total iterations), using reconstruction loss and slot regularization to establish a diverse and meaningful initial assignment of slots.
% This is followed by a slot stabilization phase using only reconstruction loss to stabilize on their initial assignments.
% Finally, in the mutual refinement phase~(20-100\%), we activate our contrastive objectives alongside the reconstruction loss to progressively calibrate the semantic boundaries at a fine-grained level.

% ================================================= %
\subsection{Comparison to State-of-the-art Methods}
In Tab.~\ref{tab:results}, we evaluate SRL against state-of-the-art methods for video object-centric learning and observe consistent gains. 
As observed, our SRL improves reproduced SlotContrast$^\dagger$~\citep{slotcontrast} by 5.5\%~(FG-ARI) and 8.8\%~(mBO) on MOVi-C.
In addition, on the real-world YTVIS dataset, our method is even more effective, enhancing SlotContrast$^\dagger$ by 18.5\%~(FG-ARI) and 8.2\%~(mBO).
These results validate that SRL enhances FG-ARI by (1) promoting clear semantic boundaries and (2) encouraging one-to-one slot-object assignments, while mBO is also greatly improved by deblurring the object boundaries.
% Our model's comparatively lower mBO on synthetic datasets stems from VideoSAUR's specialized, motion-centric objective. 
On synthetic datasets, despite our superior performance on FG-ARI, our model achieves a comparatively lower mBO score than VideoSAUR~\citep{videosaur}. 
We attribute this to VideoSAUR's specialized, motion-centric training objective, which is well-aligned with the primary characteristics of these datasets.
The objects in the MOVi datasets exhibit highly constrained degrees of freedom; they are non-deformable, and their motion is restricted to rigid transformations such as translation and rotation. 
Thus, VideoSAUR's learning process is tailored to such monotonous scenario, which excels at grouping patches with consistent motion. 
% This creates an ideal scenario for VideoSAUR's learning process, which excels at grouping patches with consistent motion.
This directly translates to more precise boundary segmentation and, consequently, a higher mBO score in this controlled environment.
However, our model's strong performance on the more challenging YTVIS, which features objects with higher degrees of freedom and non-rigid deformations, demonstrates the effectiveness of SRL in capturing diverse and complex object boundaries, suggesting greater generalizability to real-world scenarios. 
Qualitative results are in the Appendix.

\subsection{Object Dynamics Prediction}
To test whether our method benefits downstream tasks, we evaluate our pretrained video object-centric models on an object dynamics prediction task. Following SlotContrast, we attach a dynamic module on top of the frozen object-centric encoder and train it to predict future slots. We adopt SlotFormer~\citep{slotformer} for the dynamic module, which performs multiple rollout steps to infer the dynamics of object slots after a set of burn-in frames.

For a fair comparison, we use the identical experimental setup introduced in SlotContrast. Experiments are conducted on MOVi-C, MOVi-E, and YTVIS-2021, and results are reported in Tab.~\ref{tab:supple_object_dynamics}.
Across all datasets, our method consistently outperforms both the reconstruction-only baseline and SlotContrast, demonstrating that SRL yields object-centric features more amenable to modeling temporal evolution. 
Note that the performance on MOVi-E is nearly saturated, so the results do not differ significantly across methods.
These results suggest that SRL not only improves static object discovery but also produces representations that better capture object dynamics in realistic video settings.

\begin{table}[t]
\centering
\small
\vspace{-10pt}
\caption{Experimental results on object dynamics prediction. Predictions are obtained by integrating each frozen pretrained model into the SlotFormer~(SF) framework.}
\label{tab:supple_object_dynamics}
\vspace{-5pt}
\renewcommand{\arraystretch}{0.92}  % Default value: 1
\setlength{\tabcolsep}{9.5pt} % Default value: 6pt
\begin{tabular}{l cc cc cc}
\toprule
\multirow{2}{*}{\textbf{Method}} &
\multicolumn{2}{c}{\textbf{MOVi‑C}} &
\multicolumn{2}{c}{\textbf{MOVi‑E}} & \multicolumn{2}{c}{\textbf{YouTube-VIS}} \\
\cmidrule(lr){2-3}\cmidrule(lr){4-5}\cmidrule(lr){6-7}
& FG‑ARI\,$\uparrow$ & mBO\,$\uparrow$
& FG‑ARI\,$\uparrow$ & mBO\,$\uparrow$
& FG‑ARI\,$\uparrow$ & mBO\,$\uparrow$ \\ 
\midrule
Reconstruction + SF & 50.7 & 25.9 & \textbf{70.6} & 24.3 & 27.4 & 28.9 \\ 
SlotContrast + SF & 63.8 & 26.1 & 70.5 & \textbf{24.9} & 29.2 & 29.6 \\
\rowcolor{gray!20}
SRL~(Ours) + SF & \textbf{68.9} & \textbf{27.4} & 70.4 & \textbf{24.9} & \textbf{32.2} & \textbf{30.0} \\ %
\bottomrule
\end{tabular}
\vspace{-10pt}
\end{table}

\subsection{Ablation Study}
% component analysis
% enc top-k
% slot uniform M <- half일 경우 1/4, 1/3 으로 실험
All studies are conducted on the MOVi-C dataset.

\noindent\textbf{Component Ablation}
In Tab.~\ref{tab:ablation}, we examine the impact of each component, using our re-implemented SlotContrast as the baseline.
% Our baseline is re-implemented SlotContrast.
% , which achieves an FG-ARI of 70.8 and mBO of 31.4 on MOVi-C.
Introducing the decoder deblurring objective~($\mathcal{L}^{\text{CL-dec}}$) provides a substantial boost in mBO, increasing it to 33.2. 
This result validates the objective's mechanism: by explicitly penalizing ambiguity at object boundaries, it compels the decoder to produce sharper, more precise segmentation masks.
This enhanced boundary accuracy leads to a higher IoU with the ground truth, which is the basis of the mBO metric.
Conversely, activating the encoder denoising objective~($\mathcal{L}^{\text{CL-enc}}$) yields a notable improvement in FG-ARI.
By aligning the noisy patches correctly, the model achieves a more coherent and temporally stable clustering of foreground pixels.
% \begin{table}[t]
% \centering
% % \small
% \caption{Component ablation. \textcolor{blue}{이거 그냥 movi-c만 남기기..? 분량도 많으니}}
% \label{tab:ablation}
% \renewcommand{\arraystretch}{1.}  % Default value: 1
% \setlength{\tabcolsep}{6pt} % Default value: 6pt
% \begin{tabular}{ccc  cc cc cc}
% \toprule
% \multicolumn{2}{c}{SRL} & \multirow{2}{*}{Reg} & \multicolumn{2}{c}{\textbf{MOVi‑C}} &
% \multicolumn{2}{c}{\textbf{MOVi‑E}} & \multicolumn{2}{c}{\textbf{YouTube-VIS}} \\
% \cmidrule(lr){1-2}\cmidrule(lr){4-5}\cmidrule(lr){6-7}\cmidrule(lr){8-9}
% Enc & Dec &
% & FG‑ARI\,$\uparrow$ & mBO\,$\uparrow$
% & FG‑ARI\,$\uparrow$ & mBO\,$\uparrow$
% & FG‑ARI\,$\uparrow$ & mBO\,$\uparrow$ \\ 
% \midrule
%  &  &  & 70.8 & 31.4 &  &  & 35.2 & 32.7 \\ %
% \checkmark &  &  & 72.2 & 31.2 &  &  &  37.0 & 33.8 \\ % 15만iter: 71.5, 33.0
%  & \checkmark &  & 70.0 & 33.2 &  &  & 37.9 & 33.2 \\ % 15만iter: 71.8, 34.0
% \checkmark & \checkmark &  & 70.7 & 35.1 &  &  &  37.4 & 34.0 \\ % 15만iter: 71.8, 32.6
% \checkmark & & \checkmark & 74.2 & 33.2 &  &  &  41.0 & 35.4 \\ %
%  & \checkmark & \checkmark & 73.0 & 33.5 &  &  &  43.7 & 37.3 \\ %
%  & & \checkmark & 72.8 & 32.3 &  &  & 38.4 & 33.1 \\ %
% \checkmark & \checkmark & \checkmark & 74.7 & 34.8 &  &  & 43.7 & 36.8  \\ %
% \bottomrule
% \end{tabular}
% \end{table}

\begin{wrapfigure}{r}{0.40\textwidth}
    \centering
    % \vspace{-64pt}
    \vspace{-13pt}
    \small
    \captionof{table}{Component ablation study.}
    \vspace{-10pt}
    \label{tab:ablation}
    \renewcommand{\arraystretch}{0.89}  % Default value: 1
    \setlength{\tabcolsep}{2.4pt} % Default value: 6pt
        \begin{tabular}{ccc  cc}
            \toprule
            \multirow{1}{*}{Deblur} & \multirow{1}{*}{Denoise} & \multirow{1}{*}{Reg} & \multicolumn{2}{c}{\textbf{MOVi‑C}} \\
            \cmidrule(lr){4-5}
            $\mathcal{L}^{\text{CL-dec}}$& $\mathcal{L}^{\text{CL-enc}}$& $\mathcal{L}^{\text{reg}}$& FG‑ARI\,$\uparrow$ & mBO\,$\uparrow$ \\ 
            \midrule
             &  &  & 70.8 & 31.4 \\ %
            \checkmark &  &  & 70.0 & 33.2 \\ % 15만iter: 71.8, 34.0
             & \checkmark &  & 72.2 & 31.2 \\ % 15만iter: 71.5, 33.0
            \checkmark & \checkmark &  & 70.7 & 35.1 \\ % 15만iter: 71.8, 32.6
            \checkmark &  & \checkmark & 73.0 & 33.5 \\ %
             & \checkmark & \checkmark & 74.2 & 33.2 \\ %
             % & & \checkmark & 72.8 & 32.3 \\ %
             \rowcolor{gray!20}
            \checkmark & \checkmark & \checkmark & 74.3 & 34.5  \\ %
            % \checkmark & \checkmark & \checkmark & 74.7 & 34.8  \\ %
            \bottomrule
        \end{tabular}
        \vspace{-16pt}
\end{wrapfigure}
% Conversely, activating the Decoder Deblurring objective~($\mathcal{L}^{\text{CL-dec}}$) provides a substantial boost in mBO, increasing it to 33.2. 
% This result directly validates the objective's mechanism: by explicitly penalizing ambiguity at object boundaries, it compels the decoder to produce sharper, more precise segmentation masks. 
% This enhanced boundary accuracy leads to a higher Intersection-over-Union (IoU) with the ground truth, which is the basis of the mBO metric.
Crucially, the full synergistic potential of our SRL is unlocked when they are built upon the foundation laid by our slot regularization.
This initial regularization establishes a well-differentiated semantic space by minimizing the overlap between slot representations.
By ensuring that each slot is initialized with a distinct object-level concept, we prevent the subsequent denoising and deblurring objectives from operating on fragmented representations where the model would inadvertently learn to sharpen the semantic boundaries between object fragments.
% rather than correcting the initial over-fragmentation.

\noindent\textbf{Effectiveness of Hierarchy in Contrastive Learning}
To validate the necessity of our hierarchical design, we compare its results against two simplified, single-level variants in Tab.~\ref{tab:ablation_contrastive}; the second row uses only the primary positive set~($\mathcal{P}$) and treats all other patches, including semi-positives, as negatives, while the third row uses only the semi-positive set~($\mathcal{Q}$) as the sole source of positive signal.

\begin{wrapfigure}{r}{0.40\textwidth}
    \centering
    \vspace{-13pt}
    \small
    \captionof{table}{Ablation study of hierarchical contrastive objective.
    Pos., S.Pos., and Time indicate whether the positive set $\mathcal{P}$, the semi-positive set $\mathcal{Q}$, and the temporal sampling strategy are used or not.}
    \vspace{-6pt}
    \label{tab:ablation_contrastive}
    \renewcommand{\arraystretch}{0.8}  % Default value: 1
    \setlength{\tabcolsep}{4pt} % Default value: 6pt
        \begin{tabular}{ccc  cc}
            \toprule
            \multirow{2}{*}{Pos.} & \multirow{2}{*}{S.Pos.} & \multirow{2}{*}{Time} & \multicolumn{2}{c}{\textbf{MOVi‑C}} \\
            \cmidrule(lr){4-5}
             &  &  & FG‑ARI\,$\uparrow$ & mBO\,$\uparrow$ \\ 
            \midrule
            \rowcolor{gray!20}
            \checkmark & \checkmark & \checkmark & 74.3 & 34.5  \\ %
            % \midrule
            \checkmark &  & \checkmark & 67.2 & 32.4  \\ %
             & \checkmark & \checkmark & 69.9 & 32.7  \\ %
             % \midrule
            \checkmark & \checkmark &  & 72.0 & 34.4  \\ %
            \bottomrule
        \end{tabular}
        \vspace{-12pt}
\end{wrapfigure}

Both simplifications lead to a degradation in performance, but for different reasons. 
The positive-only variant suffers from a severe false negative problem; it incorrectly penalizes patches that belong to the same object but are not among the highest-confidence anchors~(e.g., not Top-$K$ similar for the encoder, nor the anchor itself for the decoder). 
This corrupts the semantic space and leads to fragmented representations. 
The semi-positive-only variant is also suboptimal, as it forces one module to exclusively mimic the other's potentially flawed representation without a stable grounding signal. 
For instance, it would compel the decoder to perfectly replicate the encoder's sharp but noisy groupings, preventing it from learning a more spatially coherent mask.
These results confirm the necessity of our hierarchical structure.

\noindent\textbf{Importance of Temporal Context in Contrastive Sampling}
Our framework gathers positive and semi-positive candidates from all $T$ frames available in a video clip. 
To investigate the importance of this temporal context, we conduct an ablation where contrastive sets are sourced exclusively from an anchor's current frame.
The results in Tab.~\ref{tab:ablation_contrastive}~(last row) reveal that the impact on mBO is marginal since the blurring effect is primarily the spatial phenomenon, yet the semantic clustering~(FG-ARI) benefits immensely from temporal context.
Therefore, to achieve robust and temporally-consistent predictions in videos, we claim that it is crucial to leverage a temporal window.

\begin{figure}[t]
\vspace{-15pt}
  \centering
  \begin{subfigure}{0.24\linewidth}
    % \fbox{\rule{0pt}{0.5in} \rule{.9\linewidth}{0pt}}
        \includegraphics[width=1\textwidth]{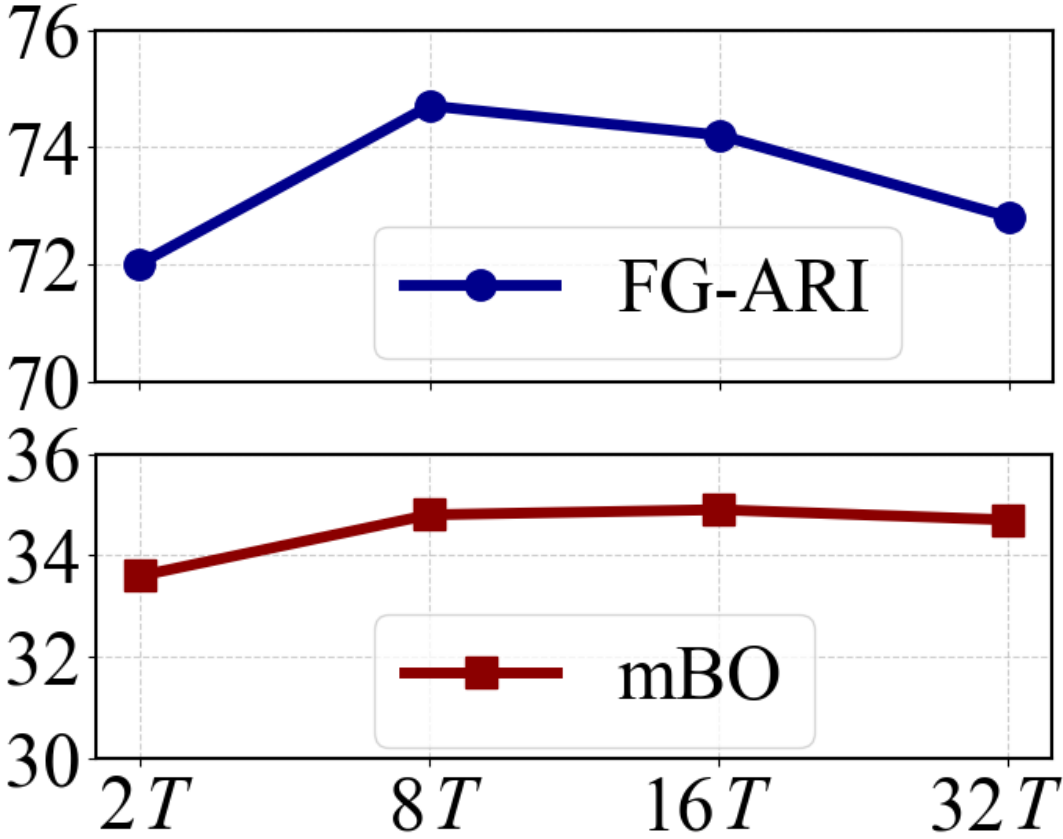}
    \vspace{-15pt}
    \caption{$K$}
    \label{fig.hyperparameter_a}
  \end{subfigure}
  \hfill
  \begin{subfigure}{0.24\linewidth}
    % \fbox{\rule{0pt}{0.5in} \rule{.9\linewidth}{0pt}}
    \includegraphics[width=1\textwidth]{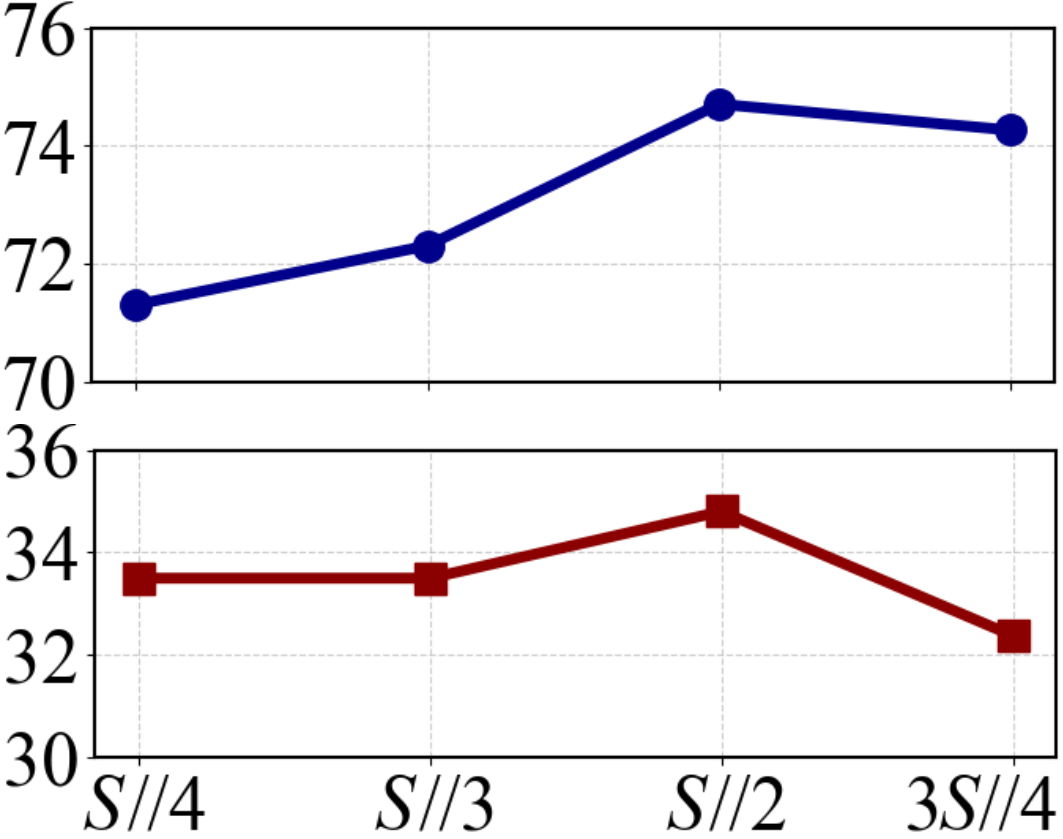}
    \vspace{-15pt}
    \caption{$M$}
    \label{fig.hyperparameter_b}
  \end{subfigure}
    \hfill
  \begin{subfigure}{0.24\linewidth}
    % \fbox{\rule{0pt}{0.5in} \rule{.9\linewidth}{0pt}}
    \includegraphics[width=1\textwidth]{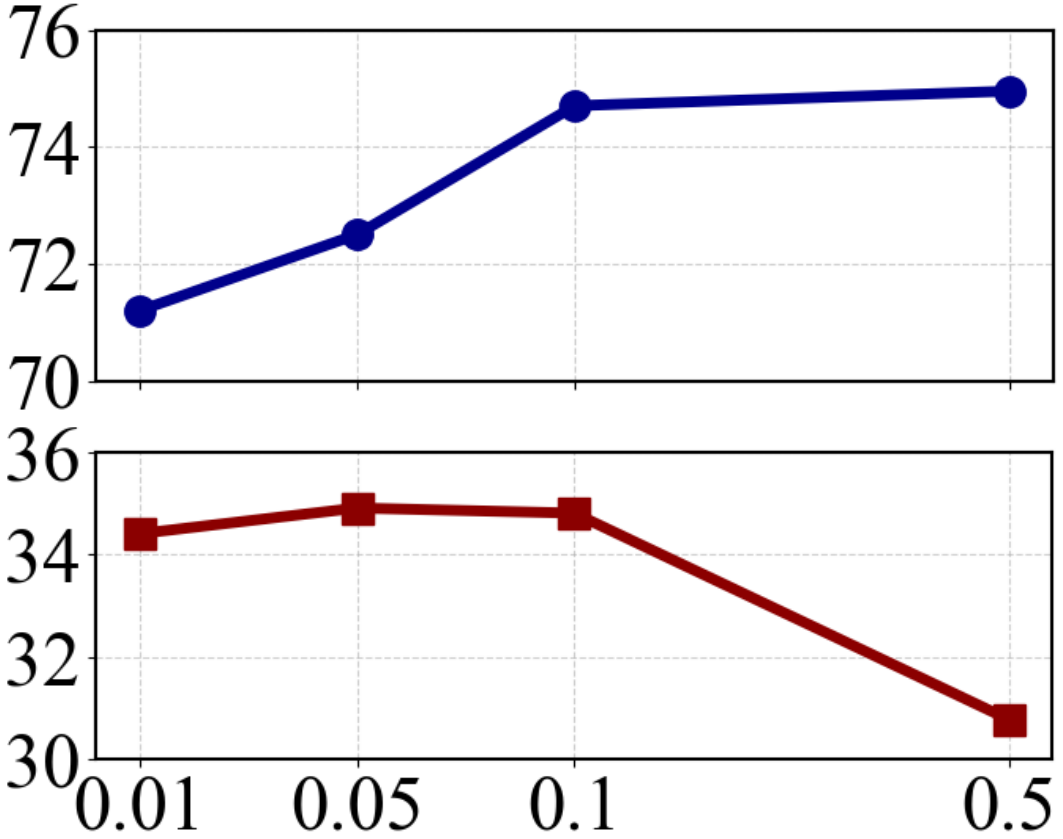}
    \vspace{-15pt}
    \caption{$\lambda^\text{reg}$}
    \label{fig.hyperparameter_c}
    \end{subfigure}
    \hfill
  \begin{subfigure}{0.24\linewidth}
    % \fbox{\rule{0pt}{0.5in} \rule{.9\linewidth}{0pt}}
    \includegraphics[width=1\textwidth]{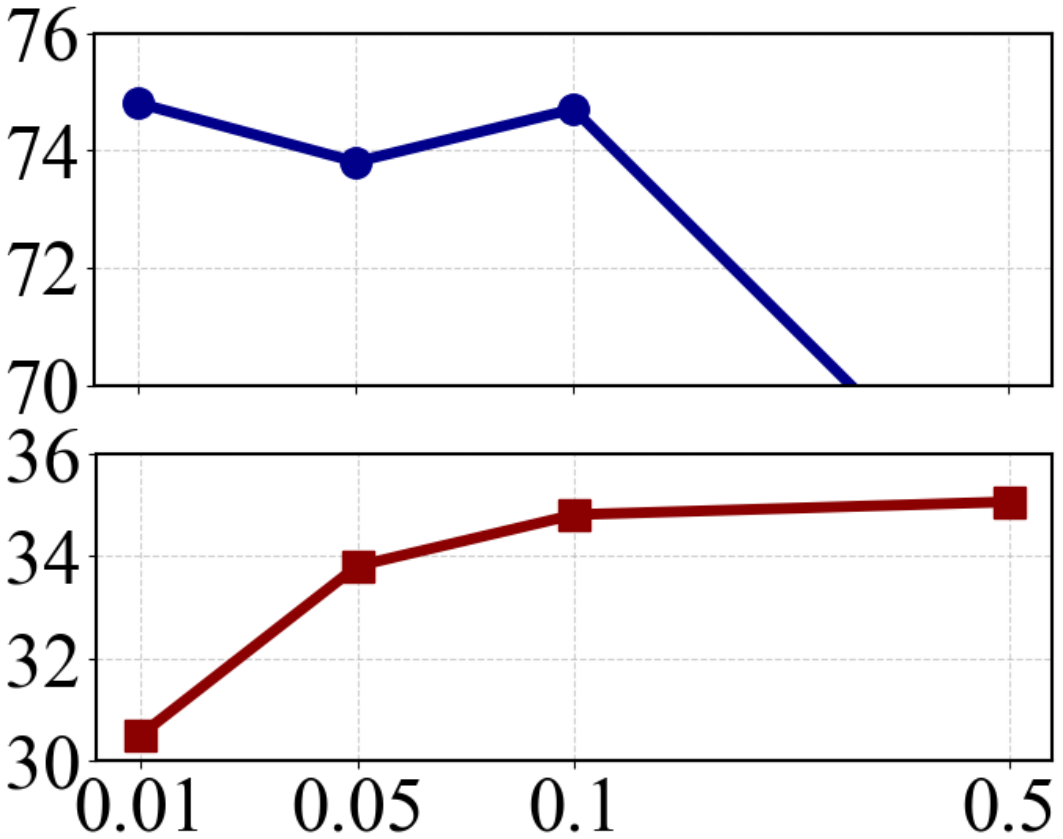}
    \vspace{-15pt}
    \caption{$\lambda^\text{CL}$}
    \label{fig.hyperparameter_d}
  \end{subfigure}
  \vspace{-10pt}
  \caption{
    Ablation study on coefficients. For (c), we vary $\lambda^{\text{reg}}$ with fixed $\lambda^{\text{CL}}$, and vice versa for (d).
    }
  \label{fig_hyperparameter}
  \vspace{-15pt}
\end{figure}

\noindent\textbf{Number of positive patches for denoising contrastive learning $K$}
In Fig.~\ref{fig.hyperparameter_a}, We study the sensitivity of our denoising contrastive learning module to the number of positive neighbors, $K$, used in the positive set $\mathcal{P}^{\text{enc}}$.
On MOVi-C, the best results occur around $K = 8T$, striking a balance between semantic coverage and noise.
Importantly, SRL is robust to the choice of $K$: performance fluctuates only marginally and consistently outperforms SlotContrast across a wide range of $K$.
% We study the sensitivity of our denoising contrastive learning module to the number of positive neighbors, $K$, used in the positive set $\mathcal{P}_i^{\text{enc}}$.
% The results indicate that performance is optimal at $K=8T$ on MOVi-C, as it strikes a balance between capturing sufficient semantic diversity and preventing the inclusion of less relevant samples.
% Yet, we claim that SRL is robust to $K$ as 성능이 크게 안벼해 (FG-ARI +- 0.0 mBO +- 0.0)

\noindent\textbf{Number of penalized slots $M$}
We also analyze the impact of $M$, the number of redundant slots penalized by our slot regularization during the warm-up phase, in Fig.~\ref{fig.hyperparameter_b}.
This parameter determines how aggressively the model prunes overlapping slot assignments. 
Our analysis reveals that a simple yet effective heuristic, setting $M$ to half the total number of slots~($\lfloor S/2 \rfloor$), consistently achieves decent performances robustly preventing slot collapse.
Thus, $M$ is set to $\lfloor S/2 \rfloor$ across all datasets.

\noindent\textbf{Loss coefficients}
We analyze the sensitivity of the loss coefficients $\lambda^{\text{reg}}$ and $\lambda^{\text{CL}}$, in Fig.~\ref{fig.hyperparameter_c} and Fig.~\ref{fig.hyperparameter_d}.
Results are reasonably stable near the default, but extreme values degrade performance.
Increasing $\lambda^{\text{reg}}$ promotes slot uniformity~(discouraging semantic overlap) yet acts as a smoothing prior that can blur boundaries and lower mBO; decreasing it too much under-constrains slots, inducing over-fragmentation and reducing FG-ARI.
Increasing $\lambda^{\text{CL}}$ strengthens local discrimination and edge sharpening, but when set too high, it over-separates fine-grained features, lowering FG-ARI.
On the other hand, making it too small lets reconstruction dominate~(a low-pass effect), weakening edge cues and lowering mBO.
Still, the hyperparameter choice is straightforward: we use a single fixed setting $\lambda^{\text{reg}}=\lambda^{\text{CL}}=0.1$, which performs reliably across all datasets.
% We performed a sensitivity analysis on the loss coefficients $\lambda^{\text{reg}}$ and $\lambda^{\text{CL}}$.
% Accordingly, we adopt a single configuration $\lambda^{\text{reg}}=\lambda^{\text{CL}}=0.1$ for all benchmarks, which works reliably without dataset-specific tuning.
% The model demonstrates robustness to minor variations in these weights, indicating that it does not require extensive, dataset-specific hyperparameter tuning.
% For all reported results, we used a single, fixed setting of $\lambda^{\text{reg}} = \lambda^{\text{CL}} = 0.1$, which proved consistently effective across all benchmarks.

\subsection{Effectiveness of synergistic learning between Attn and Mask}

\begin{wrapfigure}{r}{0.53\textwidth}
\centering
\vspace{-15pt}
\includegraphics[width=\linewidth, trim=0.2 0 0.2 0, clip]{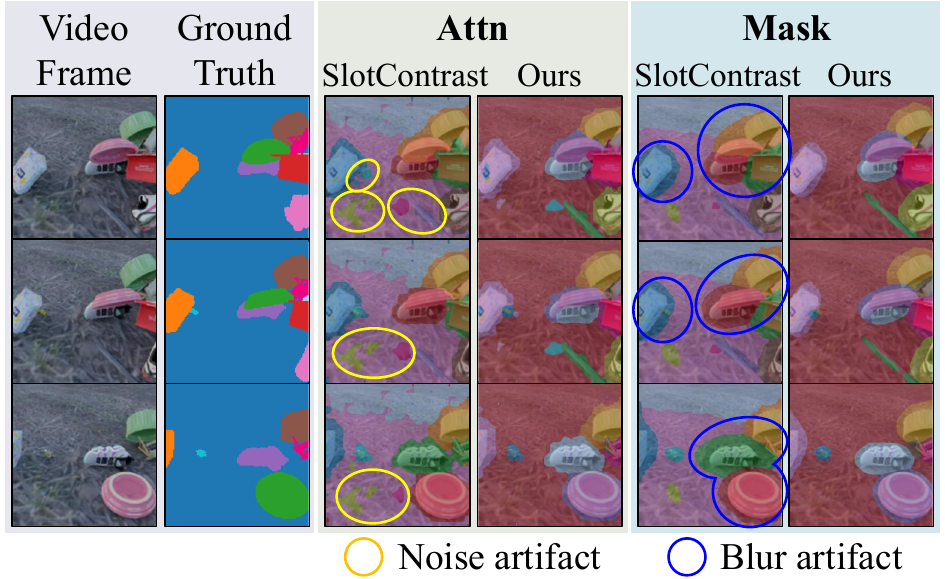}
\vspace{-20pt}
\caption{Visualization of \textbf{Attn} and \textbf{Mask}.}
\label{fig.visualization_attn_mask}
\vspace{-15pt}
\end{wrapfigure}
We qualitatively compare two distinct spatial maps, the slot attention maps \textbf{Attn} and decoder predictions \textbf{Mask}, on the MOVi-C dataset between SlotContrast~\citep{slotcontrast} and our method in Fig.~\ref{fig.visualization_attn_mask}.
As observed, SlotContrast frequently yields noisy \textbf{Attn} maps, which, when coupled with the decoder's blurry \textbf{Mask}, deteriorate a vicious cycle and lead to inconsistent and noisy slot representations.
In contrast, our approach extends SlotContrast~\citep{slotcontrast} by introducing synergistic objectives that facilitate mutual refinement between the \textbf{Attn} and \textbf{Mask} representations.
This process leverages their complementary strengths, producing denoised and deblurred predictions. 
As a result, the two spatial maps become more consistent with one another, demonstrating the effectiveness of our synergistic learning framework.

\subsection{Slot Specialization}

\begin{wrapfigure}{r}{0.35\textwidth}
\centering
\vspace{-12pt}
\includegraphics[width=\linewidth, trim=0.2 0 0.2 0, clip]{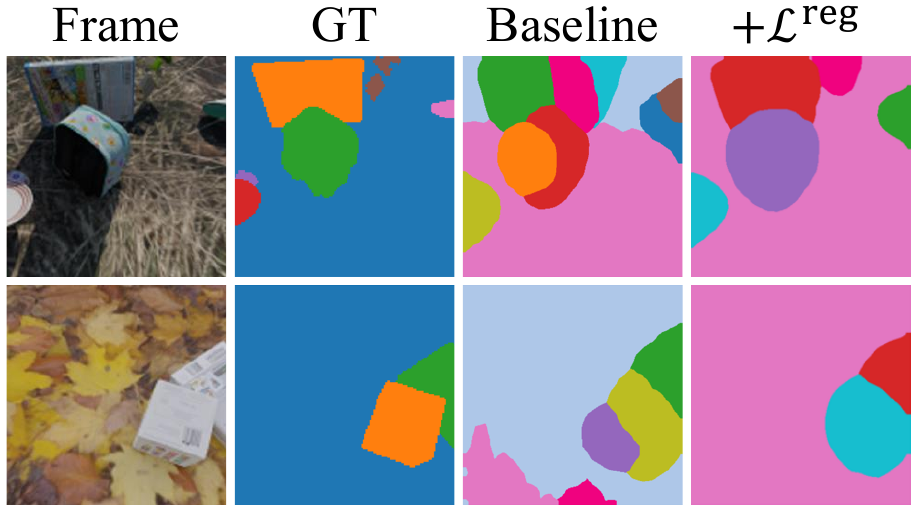}
\vspace{-15pt}
\caption{Visualization of decoder's final prediction, \textbf{Mask}.}
\vspace{-10pt}
\label{fig.exp_slot_specialization}
\end{wrapfigure}
To discourage multiple slots from redundantly capturing the same object representation, we introduce slot regularization during the warm-up stage of training. 
We assess its impact by visualizing predicted masks on the MOVi-C dataset, comparing the baseline with and without $\mathcal{L}^{\text{reg}}$~(Fig.~\ref{fig.exp_slot_specialization}). 
The visualization demonstrates that slot regularization reduces object over-fragmentation by encouraging greater disparity among slots that would otherwise collapse onto the same semantics. 
This promotes a more effective one-to-one correspondence between slots and objects, thereby strengthening the synergy of our overall representation learning framework.

\section{Conclusion}
We presented a novel framework that addresses a critical, previously overlooked bottleneck in video object-centric learning: the representational conflict between the encoder's sharp but noisy groupings and the decoder's coherent but blurry reconstructions. 
Our solution, Synergistic Representation Learning, introduces two purpose-built, ternary contrastive objectives that allow the encoder and decoder to enter a virtuous cycle of mutual refinement. 
The effectiveness of our approach, validated by state-of-the-art performance, demonstrates that explicitly modeling and resolving the discrepancies between a model's internal representations is a powerful mechanism for enhancing performance.

%%

% \centerline{\bf Numbers and Arrays}
% \bgroup
% \def\arraystretch{1.5}
% \begin{tabular}{p{1in}p{3.25in}}
% $\displaystyle a$ & A scalar (integer or real)\\
% $\displaystyle \va$ & A vector\\
% $\displaystyle \mA$ & A matrix\\
% $\displaystyle \tA$ & A tensor\\
% $\displaystyle \mI_n$ & Identity matrix with $n$ rows and $n$ columns\\
% $\displaystyle \mI$ & Identity matrix with dimensionality implied by context\\
% $\displaystyle \ve^{(i)}$ & Standard basis vector $[0,\dots,0,1,0,\dots,0]$ with a 1 at position $i$\\
% $\displaystyle \text{diag}(\va)$ & A square, diagonal matrix with diagonal entries given by $\va$\\
% $\displaystyle \ra$ & A scalar random variable\\
% $\displaystyle \rva$ & A vector-valued random variable\\
% $\displaystyle \rmA$ & A matrix-valued random variable\\
% \end{tabular}
% \egroup
% \vspace{0.25cm}

% \centerline{\bf Sets and Graphs}
% \bgroup
% \def\arraystretch{1.5}

% \begin{tabular}{p{1.25in}p{3.25in}}
% $\displaystyle \sA$ & A set\\
% $\displaystyle \R$ & The set of real numbers \\
% $\displaystyle \{0, 1\}$ & The set containing 0 and 1 \\
% $\displaystyle \{0, 1, \dots, n \}$ & The set of all integers between $0$ and $n$\\
% $\displaystyle [a, b]$ & The real interval including $a$ and $b$\\
% $\displaystyle (a, b]$ & The real interval excluding $a$ but including $b$\\
% $\displaystyle \sA \backslash \sB$ & Set subtraction, i.e., the set containing the elements of $\sA$ that are not in $\sB$\\
% $\displaystyle \gG$ & A graph\\
% $\displaystyle \parents_\gG(\ervx_i)$ & The parents of $\ervx_i$ in $\gG$
% \end{tabular}
% \vspace{0.25cm}

\bibliography{iclr2026_conference}
\bibliographystyle{iclr2026_conference}

\appendix

\renewcommand{\thesection}{A}
\renewcommand{\thetable}{A\arabic{table}}   
\renewcommand{\thefigure}{A\arabic{figure}}
\setcounter{section}{0}
\setcounter{table}{0}
\setcounter{figure}{0}

%%%%%%%%%%%%%%%%%%%%%%%%%%%%%%%%%%%%%%
\section{Visualization}
%%%%%%%%%%%%%%%%%%%%%%%%%%%%%%%%%%%%%%

\begin{figure*}[h]
\centering
\includegraphics[width=0.98\columnwidth]{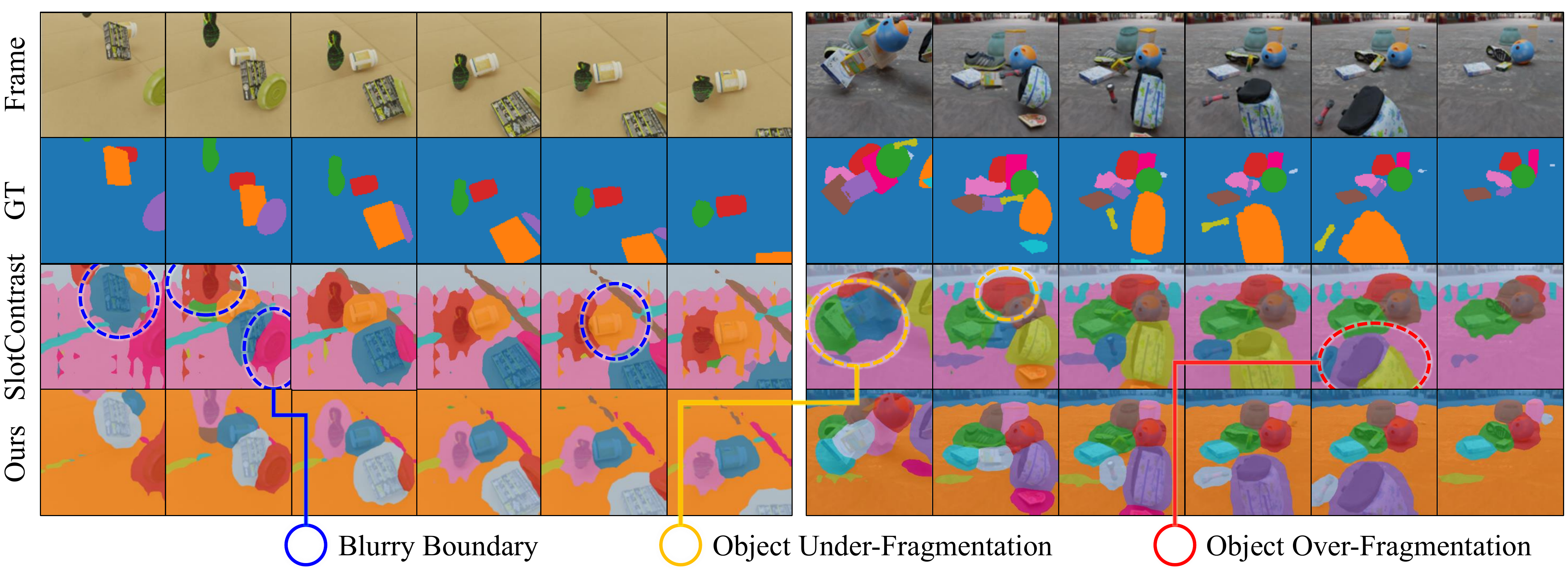} 
\caption{Qualitative comparison results of ours and SlotContrast on MOVi-C dataset.
}
\label{fig1.qualitative.movic}
\end{figure*}
\begin{figure*}[h]
\centering
\includegraphics[width=0.98\columnwidth]{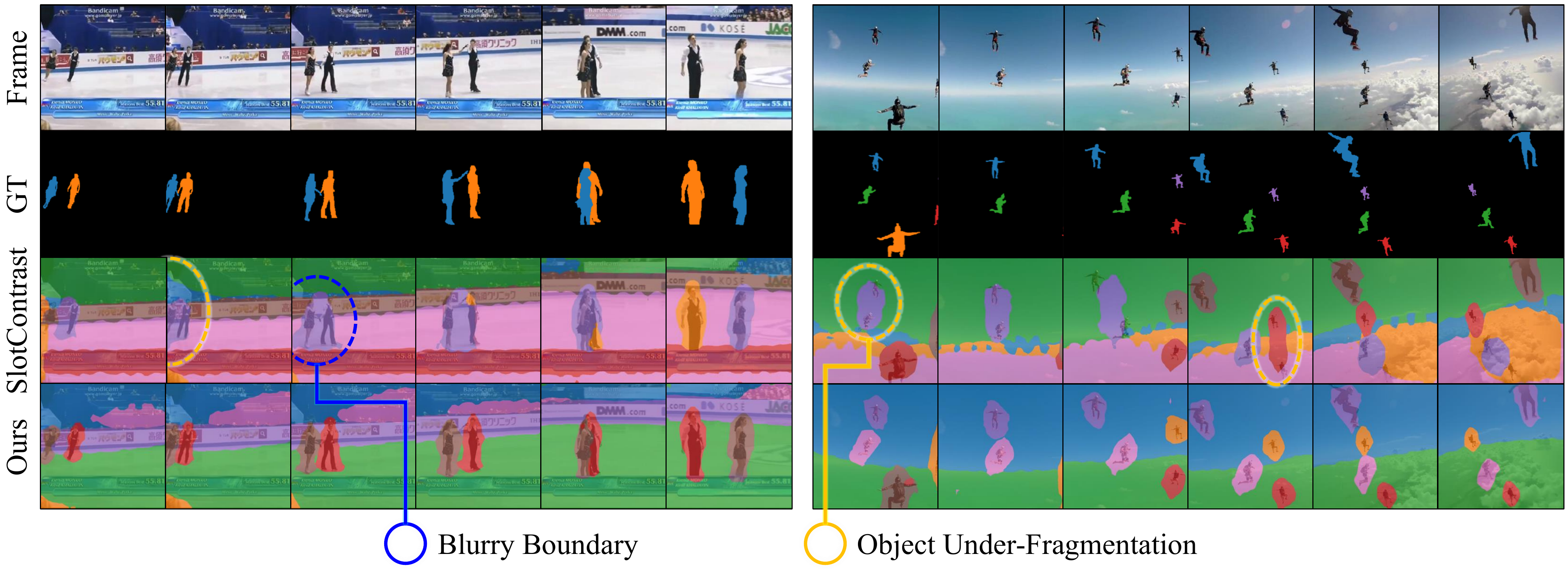} 
\caption{Qualitative comparison results of ours and SlotContrast on YTVIS 2021 dataset.
}
\label{fig1.qualitative.ytvis}
\end{figure*}

\subsection{Qualitative Results}
For qualitative evaluation, we compare our method with SlotContrast~\citep{slotcontrast} on the MOVi-C and YTVIS 2021 datasets, as shown in Fig.~\ref{fig1.qualitative.movic} and Fig.~\ref{fig1.qualitative.ytvis}. 

On the MOVi-C dataset, our method demonstrates a notable improvement in object separation. 
As illustrated in the left video example, our baseline~(SlotContrast) often produces blurry decoded masks where slots exhibit diffuse spatial support, extending beyond the object's true boundaries. 
In contrast, our method generates compact slots that adhere more faithfully to the object's contours, resulting in clearer semantic boundaries. 
Furthermore, the right example shows how these sharp boundaries directly mitigate a common failure mode of object under-fragmentation~(the erroneous grouping of multiple objects into a single slot). 
Whereas SlotContrast incorrectly merges distinct objects~(e.g., regions covered by red and green slots), our SRL framework alleviates the under-fragmentation issue by partitioning them into different slots.
Complementing this, our warm-up strategy prevents the opposing failure mode of over-fragmentation, where a single object is fragmented into different parts.
Together, these components ensure a more robust one-to-one correspondence between slots and objects.

This trend extends to the more challenging YTVIS dataset~(Fig.~\ref{fig1.qualitative.ytvis}). 
The baseline's inability to maintain compact semantic boundaries causes it to fail in scenarios with object overlap, where proximal entities are often merged into a single slot. 
For instance, in both video examples, SlotContrast incorrectly assigns one slot to cover two distinct people~(the region covered by the purple slots in both videos). 
In contrast, our method yields sharper predictions by learning to clarify the semantic boundary via denoising and deblurring contrastive objectives.
This allows slots to more faithfully specialize to individual objects, even when they overlap.

\subsection{Failure Cases}
\begin{figure*}[t]
\centering
\includegraphics[width=0.98\columnwidth]{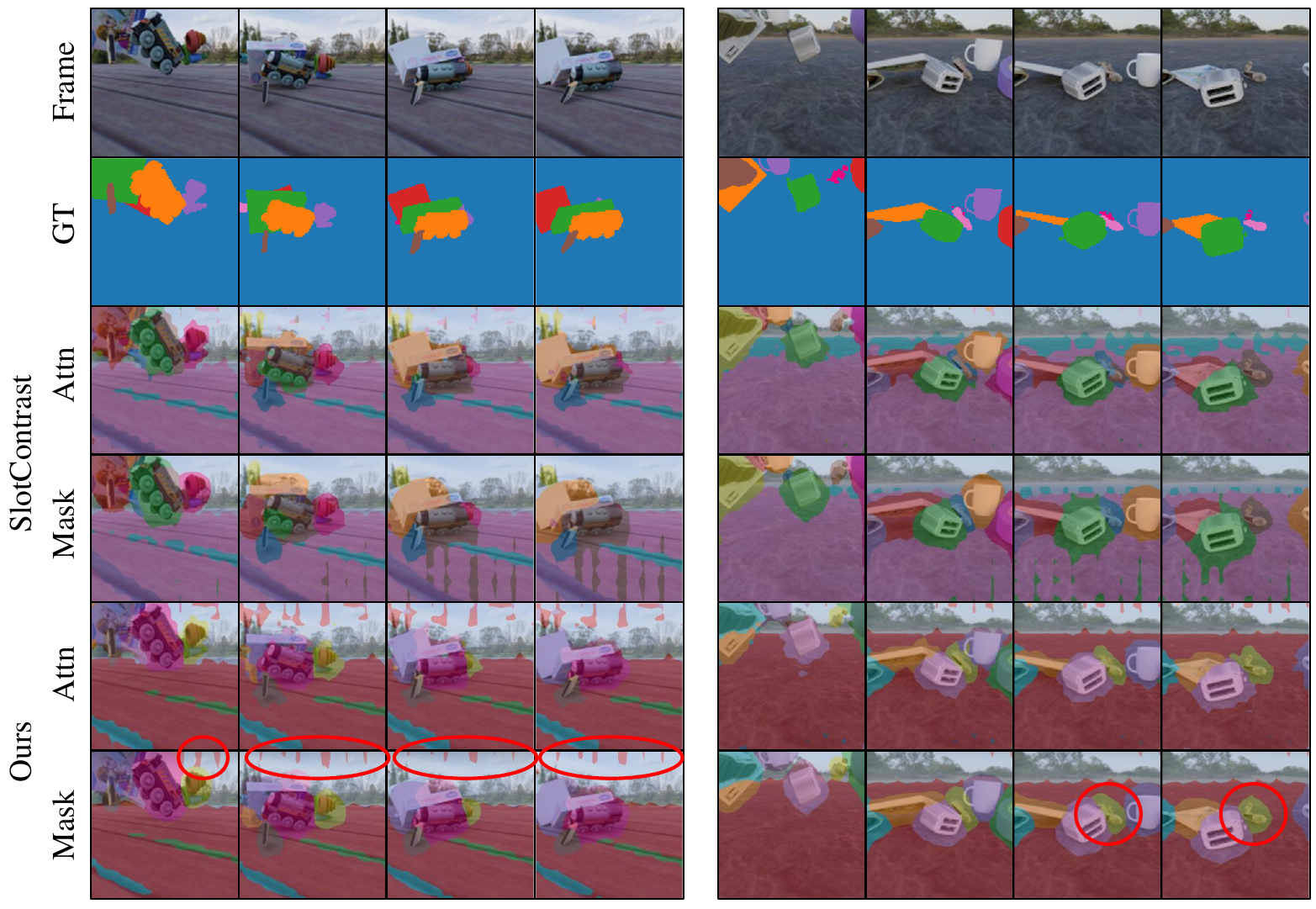} 
\caption{Failure cases on MOVi-C dataset.
}
\label{fig.supp.failure_case}
\end{figure*}

To provide a more complete analysis of our method’s behavior and limitations, we additionally visualize representative failure cases in Fig.~\ref{fig.supp.failure_case}. 
Specifically, we present two samples from the MOVi-C dataset to qualitatively examine the scenarios where our model performs suboptimally.

In this work, we focused on the discrepancy that encoder spatial maps tend to be noisy while decoder maps are overly blurry. 
However, as shown in Fig.~\ref{fig.supp.failure_case}~(left), there exist cases where the noise originating from the encoder propagates into the decoder and persists even after training~(red circles). 
This occurs because SRL is primarily designed to address the dominant issue of blurry decoded masks, and does not explicitly regularize decoder-side noise. 
As a result, certain noisy attention patterns may remain, similarly to the SlotContrast baseline. 
Although our method still alleviates over-fragmentation and reduces blur in such cases, explicitly modeling and suppressing this propagated noise remains an important direction for future work.

In addition, in Fig.~\ref{fig.supp.failure_case}~(right), we observe that our method occasionally under-fragments extremely small objects, failing to allocate a dedicated slot to each of them. 
This indicates that our model remains vulnerable when the targets are very small. 
We believe that explicitly targeting small-object discovery and representation is another promising direction for future work.

% The baseline often produces blurred masks that fail to distinguish adjacent objects, leading to under-fragmentation. 
% By introducing sharper learning objectives guided by encoder features, our approach alleviates this issue and enforces more precise object boundaries. 
% In effect, we break the vicious encoder–decoder cycle, where noisy encoder features amplify blurry decoder outputs and vice versa, and replace it with a virtuous cycle in which the two modules mutually refine each other. 
% This transformation corrects noisy representations and blurred boundaries, resulting in more accurate and stable object-centric predictions.

% For YouTube-VIS dataset, SlotContrast often produces blurry and inconsistent predictions across frames, which leads slots to miss small objects or to activate on multiple adjacent objects simultaneously. 
% Such errors further disrupt temporal consistency, causing failures in tracking the same semantics over time. 
% In contrast, our method yields sharper predictions by leveraging synergistic learning objectives. 
% This allows slots to more faithfully specialize to individual objects and to maintain consistent activation of the same object across frames.

\renewcommand{\thesection}{B}
\renewcommand{\thetable}{B\arabic{table}}
\renewcommand{\thefigure}{B\arabic{figure}}
\setcounter{table}{0}
\setcounter{figure}{0}

%%%%%%%%%%%%%%%%%%%%%%%%%%%%%%%%%%%%%%
\section{Further Experiments}
%%%%%%%%%%%%%%%%%%%%%%%%%%%%%%%%%%%%%%

%%%%%%%%%%%%%%%%%%%%%%%%%%%%%%%%%%%%%%%%%%%%%%%%%%%%%%%%%%%%%%%%%%%%%%%%%%%%%%%%
% \subsection{Slot Specialization}

% \begin{wrapfigure}{r}{0.35\textwidth}
% \centering
% \vspace{-12pt}
% \includegraphics[width=\linewidth, trim=0.2 0 0.2 0, clip]{fig/fig_5_ablation_slotuniform_v2.pdf}
% \vspace{-15pt}
% \caption{Visualization of decoder's final prediction, \textbf{Mask}.}
% \vspace{-10pt}
% \label{fig.exp_slot_specialization}
% \end{wrapfigure}
% To discourage multiple slots from redundantly capturing the same object representation, we introduce slot regularization during the warm-up stage of training. 
% We assess its impact by visualizing predicted masks on the MOVi-C dataset, comparing the baseline with and without $\mathcal{L}^{\text{reg}}$~(Fig.~\ref{fig.exp_slot_specialization}). 
% The visualization demonstrates that slot regularization reduces object over-fragmentation by encouraging greater disparity among slots that would otherwise collapse onto the same semantics. 
% This promotes a more effective one-to-one correspondence between slots and objects, thereby strengthening the synergy of our overall representation learning framework.

%%%%%%%%%%%%%%%%%%%%%%%%%%%%%%%%%%%%%%%%%%%%%%%%%%%%%%%%%%%%%%%%%%%%%%%%%%%%%%%%
\subsection{MAE Loss for Reconstruction}

\begin{wrapfigure}{r}{0.35\textwidth}
    \centering
    \vspace{-13pt}
    \small
    \captionof{table}{Experimental results using MAE loss for reconstruction.}
    \vspace{-6pt}
    \label{tab:supp.maeloss}
    \renewcommand{\arraystretch}{0.8}  % Default value: 1
    \setlength{\tabcolsep}{4pt} % Default value: 6pt
        \begin{tabular}{lcc}
            \toprule
            Method & FG‑ARI\,$\uparrow$ & mBO\,$\uparrow$ \\
            \midrule
            SlotContrast & 73.24 & 27.54 \\
            \rowcolor{gray!20}
            SRL~(Ours) & 74.57 & 34.28 \\
            \bottomrule
        \end{tabular}
        \vspace{-8pt}
\end{wrapfigure}

\begin{figure*}[h]
\centering
\includegraphics[width=0.7\columnwidth]{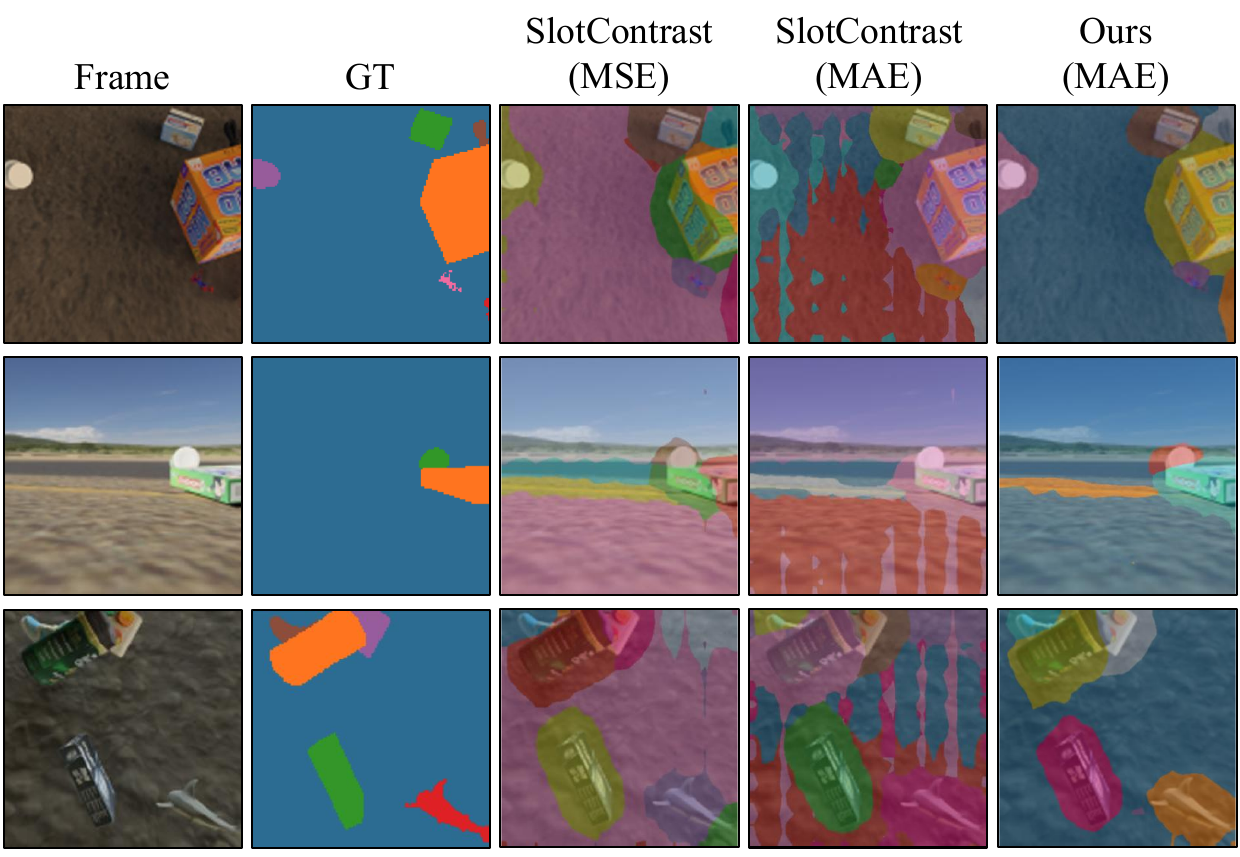} 
\caption{Qualitative comparison results when utilizing MAE loss for reconstruction objective.
}
\label{fig.supple.maeloss}
\end{figure*}

% In the main manuscript, we employed the commonly used MSE loss as the reconstruction objective. 
% To examine whether our findings hold under alternative reconstruction objectives, we additionally replace MSE with MAE loss and evaluate both SlotContrast and our SRL on MOVi-C dataset.
In the main manuscript, our SRL framework is designed to address the inherent weakness of the commonly used MSE reconstruction loss, namely, its tendency to produce blurred outputs, which causes a vicious cycle during training. To examine whether similar vulnerabilities arise under alternative reconstruction objectives, and to assess the robustness of SRL beyond the MSE setting, we additionally replace MSE with MAE loss and evaluate both SlotContrast and our SRL on MOVi-C dataset.
The quantitative and qualitative results are summarized in Tab.~\ref{tab:supp.maeloss} and Fig.~\ref{fig.supple.maeloss}, respectively.

Compared to using MSE as the reconstruction loss, MAE tends to emphasize the majority of pixels, which makes it robust to certain high error patches.
While this alleviates the over-fragmentation and improves FG-ARI,
this majority-focused behavior often causes under-fragmentation of small objects and amplifies irregular noise patterns, as shown in Fig.~\ref{fig.supple.maeloss}, which in turn degrades mBO.
% MAE poses another challenge that it poses irregular noise patterns, as shown in Fig.~\ref{fig.supple.maeloss}, which in turn degrades mBO.
Nonetheless, since our method excels at denoising such noise patterns, it is shown that SRL consistently improves the performances. 
In particular, the substantial gain in mBO indicates that our approach reduces noisy activations even when the reconstruction objective is modified from MSE to MAE.
These results demonstrate that the denoising and deblurring benefits of SRL are applicable to various challenging scenarios across different objectives~(\textit{i.e.,} MSE or MAE).

%%%%%%%%%%%%%%%%%%%%%%%%%%%%%%%%%%%%%%%%%%%%%%%%%%%%%%%%%%%%%%%%%%%%%%%%%%%%%%%%
\subsection{Experiments on Additional Datasets}
% DAVIS
% Ytvis 2019 (2019-2019, 2021-2019)
To assess whether SRL generalizes beyond the datasets used in the main manuscript, we evaluate on the DAVIS 2017~\citep{davis} validation set~($37\times37$ patch grid) using a model trained on YTVIS-2021, following the transfer protocol introduced in VideoSAUR~\citep{videosaur}. We report the boundary F-score $\mathcal{F}$ and Jaccard index $\mathcal{J}$ in Tab.~\ref{tab:supple_davis}. SRL achieves substantial improvements over SlotContrast, improving $\mathcal{J}$ by +11.7 points and the combined score $\mathcal{J}$\&$\mathcal{F}$ by +7.5 points.
% These gains indicate that our SRL discovers objects more reliably, despite never having seen DAVIS during training.

We further evaluate on YouTube-VIS 2019 (YTVIS-2019) under two scenarios:
(1) cross-dataset transfer from the model trained on YTVIS-2021, and
(2) in-dataset evaluation, where the model is trained on the YTVIS-2019 train set and evaluated on its validation set.
As summarized in Tab.~\ref{tab:supple_yt2019_from2021} and Tab.~\ref{tab:supple_yt2019_from2019}, SRL consistently surpasses SlotContrast across both settings, yielding improvements on ARI and mBO regardless of whether the model is transferred or trained in-domain.

Taken together, these results demonstrate that SRL generalizes robustly across video domains and dataset shifts, providing stronger object-centric representations than SlotContrast in both transfer and in-distribution evaluations. This suggests that the proposed learning signal not only enhances grouping quality within the training domain but also yields transferable object discovery behavior that extends to diverse video benchmarks.

\begin{figure}[h]
  \centering
  \begin{minipage}[t]{0.28\textwidth}
    \vspace{0pt}
    \centering
    \setlength{\tabcolsep}{2pt}
    \captionof{table}{Experimental results on DAVIS dataset.}
    \label{tab:supple_davis}
    \begin{tabular}{lccc}
        \toprule
        Method & $\mathcal{F}$ &  $\mathcal{J}$ &  $\mathcal{F}$\& $\mathcal{J}$ \\
        \midrule
        SlotContrast & 22.2 & 36.5 & 29.3 \\
        \rowcolor{gray!20}
        SRL~(Ours) & 25.4 & 48.2 & 36.8 \\
        \bottomrule
    \end{tabular}
  \end{minipage}\hfill
  \begin{minipage}[t]{0.31\columnwidth}
    \vspace{0pt}
    \centering
    \setlength{\tabcolsep}{1.5pt}
    \captionof{table}{YTVIS2019 results trained on YTVIS2021 dataset.}
    \label{tab:supple_yt2019_from2021}
    \begin{tabular}{lcc}
      \toprule
        Method & FG‑ARI\,$\uparrow$ & mBO\,$\uparrow$  \\
        \midrule
        SlotContrast & 16.6 & 43.3 \\
        \rowcolor{gray!20}
        SRL~(Ours) & 20.4 & 53.3 \\
        \bottomrule
    \end{tabular}
  \end{minipage}\hfill
    \begin{minipage}[t]{0.31\columnwidth}
    \vspace{0pt}
    \centering
    \setlength{\tabcolsep}{1.5pt}
    \captionof{table}{YTVIS2019 results trained on YTVIS2019 dataset.}
    \label{tab:supple_yt2019_from2019}
    \begin{tabular}{lcc}
      \toprule
        Method & FG‑ARI\,$\uparrow$ & mBO\,$\uparrow$  \\
        \midrule
        SlotContrast & 16.7	& 44.9 \\
        \rowcolor{gray!20}
        SRL~(Ours) & 19.1	& 46.9 \\
        \bottomrule
    \end{tabular}
  \end{minipage}
\end{figure}

%%%%%%%%%%%%%%%%%%%%%%%%%%%%%%%%%%%%%%%%%%%%%%%%%%%%%%%%%%%%%%%%%%%%%%%%%%%%%%%%
\subsection{Experiments on Different Pretrained Backbones}
\begin{figure}[t]
  \centering
  \begin{minipage}[t]{0.57\columnwidth}
    \vspace{0pt}
    \includegraphics[width=\linewidth]{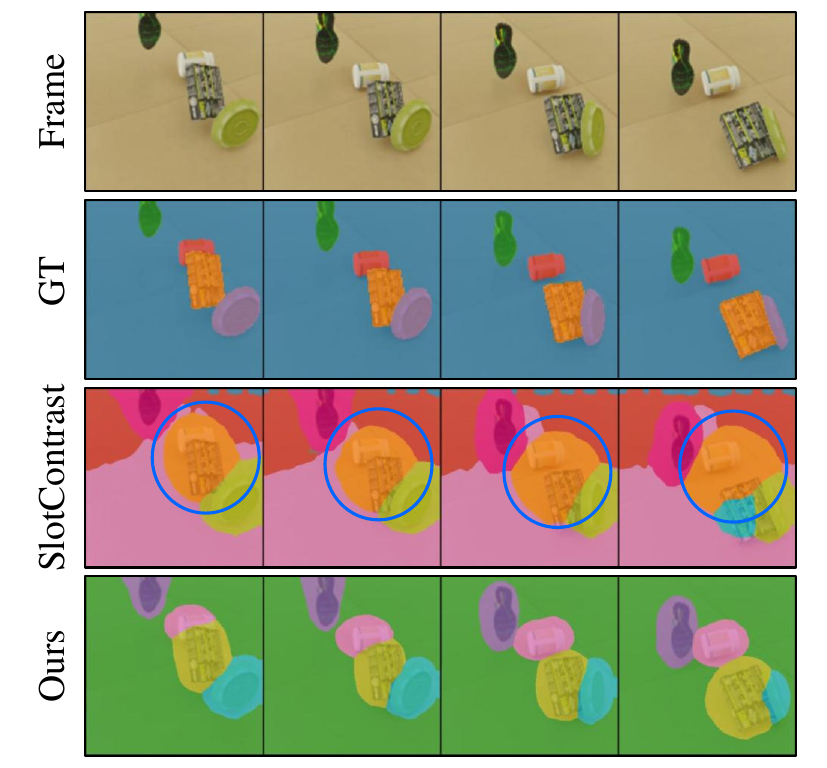}
    \vspace{-0.6cm}
    \caption{Qualitative comparison when using MoSiC as backbone encoder. Blue circles indicate position bias.}
    \label{fig:supp.mosic}
  \end{minipage}\hfill
  \begin{minipage}[t]{0.41\columnwidth}
    \vspace{0pt}
    \centering
    \setlength{\tabcolsep}{3pt}
    \captionof{table}{Experiments on MOVi-C dataset on Franca ViT-B/14.}
    \label{tab:supple_franca_movic}
    \begin{tabular}{lcc}
      \toprule
        Method & FG‑ARI\,$\uparrow$ & mBO\,$\uparrow$  \\
        \midrule
        SlotContrast & 66.8 & 35.6\\
        \rowcolor{gray!20}
        SRL~(Ours) & 66.1 & 37.2 \\
    \bottomrule
    \end{tabular}
    \vspace{0pt}
    \centering
    \setlength{\tabcolsep}{3pt}
    \captionof{table}{Experiments on YTVIS-2021 dataset on Franca ViT-B/14.}
    \label{tab:supple_franca_ytvis2021}
    \begin{tabular}{lcc}
      \toprule
        Method & FG‑ARI\,$\uparrow$ & mBO\,$\uparrow$  \\
        \midrule
        SlotContrast & 35.3 & 32.7 \\
        \rowcolor{gray!20}
        SRL~(Ours) & 38.9 & 36.4 \\
    \bottomrule
    \end{tabular}
    \vspace{0pt}
    \centering
    \setlength{\tabcolsep}{3pt}
    \captionof{table}{Experiments on MOVi-C dataset on MoSiC ViT-B/14.}
    \label{tab:supple_mosic}
    \begin{tabular}{lcc}
      \toprule
        Method & FG‑ARI\,$\uparrow$ & mBO\,$\uparrow$  \\
        \midrule
        SlotContrast & 70.3 & 31.6 \\
        \rowcolor{gray!20}
        SRL~(Ours) & 74.3 & 37.3 \\
        \bottomrule
    \end{tabular}
  \end{minipage}
\end{figure}

% MoSiC
% Neco
% DINOv2-Register
% Franca
To assess whether SRL remains effective when applied to backbone encoders beyond DINO-v2, we replace DINO-v2 with either Franca~\citep{franca} or MoSiC~\citep{mosic}, and evaluate both SlotContrast and SRL under the same training and evaluation protocol. 
The experimental results on the MOVi-C dataset are reported in Tab.~\ref{tab:supple_franca_movic}-~\ref{tab:supple_franca_ytvis2021} and Tab.~\ref{tab:supple_mosic}, respectively.

Across both backbones, SRL consistently improves performance.
When using Franca as the backbone, SRL is particularly beneficial on the mBO metric on MOVi-C, and surpasses SlotContrast by a large margin on YTVIS 2021.
SRL is also effective when built on MoSiC, a denoised backbone specifically designed to reduce feature-level noise. 
Interestingly, we observe that such denoised backbones may introduce new artifacts: MoSiC features often exhibit a strong positional bias, where slots collapse onto empty background regions or fail to track moving objects. 
This suggests that part of the denoising effect comes at the cost of distorted spatial structure~(see Fig.~\ref{fig:supp.mosic}).
Nonetheless, our method effectively mitigates these noisy artifacts and restores meaningful object assignments, thereby achieving a large performance uplift over SlotContrast. 
These results confirm that SRL not only transfers across datasets but also remains robust across diverse backbone architectures, even those subject to substantial feature-level modifications.

\subsection{Experiments on Image Data}
% COCO
\begin{wrapfigure}{r}{0.3\textwidth}
    \centering
    \vspace{-13pt}
    \small
    \captionof{table}{Results on COCO.}
    \vspace{-6pt}
    \label{tab:supp.coco}
    \renewcommand{\arraystretch}{0.8}  % Default value: 1
    \setlength{\tabcolsep}{4pt} % Default value: 6pt
        \begin{tabular}{lcc}
            \toprule
            Method & ARI\,$\uparrow$ & mBO\,$\uparrow$ \\
            \midrule
            Baseline & 40.5	& 28.8 \\
            % SlotContrast & 41.4	& 28.9 \\
            \rowcolor{gray!20}
            Ours & 42.8	& 29.4 \\
            \bottomrule
        \end{tabular}
        \vspace{-9pt}
\end{wrapfigure}

Our SRL is applicable to static images, as the conflict between the encoder's sharpness and decoder's smoothness exists in the slot attention architecture itself, independent of temporal dimensions. 
Therefore, we evaluate SRL on the MSCOCO 2017 dataset~\citep{coco} using the same training protocol as the baseline. The experiments are conducted with DINO-v2-Small/14.
As shown in Tab.~\ref{tab:supp.coco}, SRL achieves an improvement of +2.3 in ARI and +0.6 in mBO over the reconstruction-only baseline. 
This demonstrates that even without temporal cues, the mutual refinement between the encoder's sharp attention and the decoder's spatial coherence effectively improves object discovery. While our study focused on video benchmarks, these additional results confirm that SRL is a generalizable solution for object-centric learning.
% We note that SlotContrast is implemented by applying the slot contrastive objective within each single image frame.

% \textcolor{red}{확인 필요. SC는 frame 개수가 1이면 학습될게 없는거같은데}

%%%%%%%%%%%%%%%%%%%%%%%%%%%%%%%%%%%%%%%%%%%%%%%%%%%%%%%%%%%%%%%%%%%%%%%%%%%%%%%%
\subsection{Additional Ablation Study}
\begin{figure*}[t]
\centering
\includegraphics[width=0.98\columnwidth]{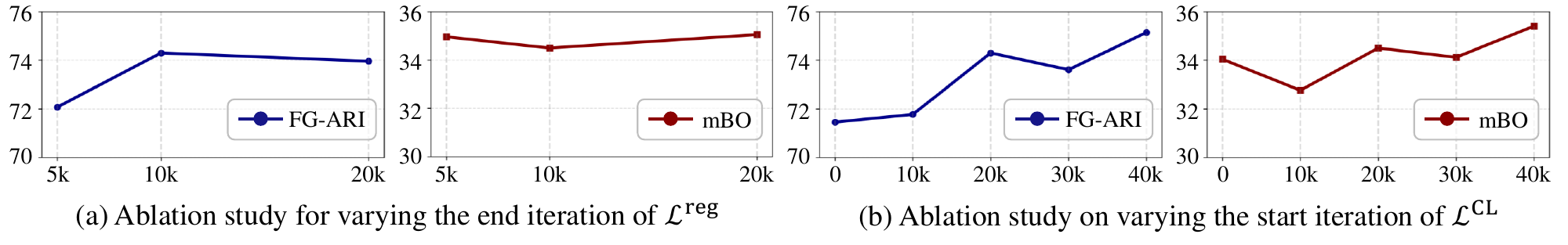} 
\caption{Ablation study on the staged training boundaries. (a) Varying the iteration at which slot regularization is turned off. (b) Varying the iteration at which contrastive learning objectives are activated.
}
\label{fig.supple.ablation_iteration}
\end{figure*}

% \eta
We conduct ablation studies to examine the robustness of our staged training strategy, which consists of:
(i) an early slot regularization phase, and (ii) a later contrastive learning phase (denoising/deblurring).
The model is trained for 100k iterations on the MOVi-C dataset, and we vary the transition point of each stage while keeping the remainder of the training settings identical. The results are summarized in Fig.~\ref{fig.supple.ablation_iteration}.

\paragraph{When to Stop Slot Regularization.}
We first vary the iteration at which slot regularization is disabled, shown in Fig.~\ref{fig.supple.ablation_iteration}~(a).
For reference, the SlotContrast baseline achieves 70.8 for FG-ARI and 31.4 for mBO on this benchmark.
Across all tested schedules, our method substantially exceeds the baseline, and the resulting performance curves remain smooth after 10k iterations. 
Stopping the regularization slightly later~(e.g., around 20k iterations) yields a modest improvement in mBO, indicating that the method does not rely on a finely tuned early cutoff. Overall, SRL maintains strong FG-ARI and mBO performance across different regularization stopping points.

\paragraph{When to Start Synergistic Representation Learning.}
Next, we vary when the contrastive learning objectives are activated during training.
The results are illustrated in Fig.~\ref{fig.supple.ablation_iteration}~(b).
Once again, all tested configurations clearly outperform SlotContrast by a significant margin. 
% Performance changes steadily as the start point is shifted earlier or later, with the best results obtained when contrastive learning is introduced in the mid-to-late training stage~(20k–40k iterations). 
We attribute this to the need for the encoder and decoder to first learn reasonably stable spatial representations; if contrastive learning is applied too early, the two branches end up guiding each other based on poorly formed features, whereas after roughly 20k iterations, the representations have largely converged and provide reliable signals. 
Nonetheless, a relatively broad starting interval still yields competitive results, indicating that the contrastive learning module is not overly sensitive to the precise activation point.
% Importantly, a relatively broad interval produces competitive results, suggesting that the contrastive learning module is not sensitive to the precise activation point.

These ablations demonstrate that SRL is robust to the choice of stage boundaries. The method consistently improves over SlotContrast under all tested schedules, and performance behaves smoothly rather than collapsing when deviating from an optimal configuration. This indicates that SRL offers a stable and reliable training procedure that does not require careful tuning of the transition point.

%%%%%%%%%%%%%%%%%%%%%%%%%%%%%%%%%%%%%%%%%%%%%%%%%%%%%%%%%%%%%%%%%%%%%%%%%%%%%%%%
\subsection{Vicious Cycle}
\begin{figure*}[h]
\centering
\includegraphics[width=0.98\columnwidth]{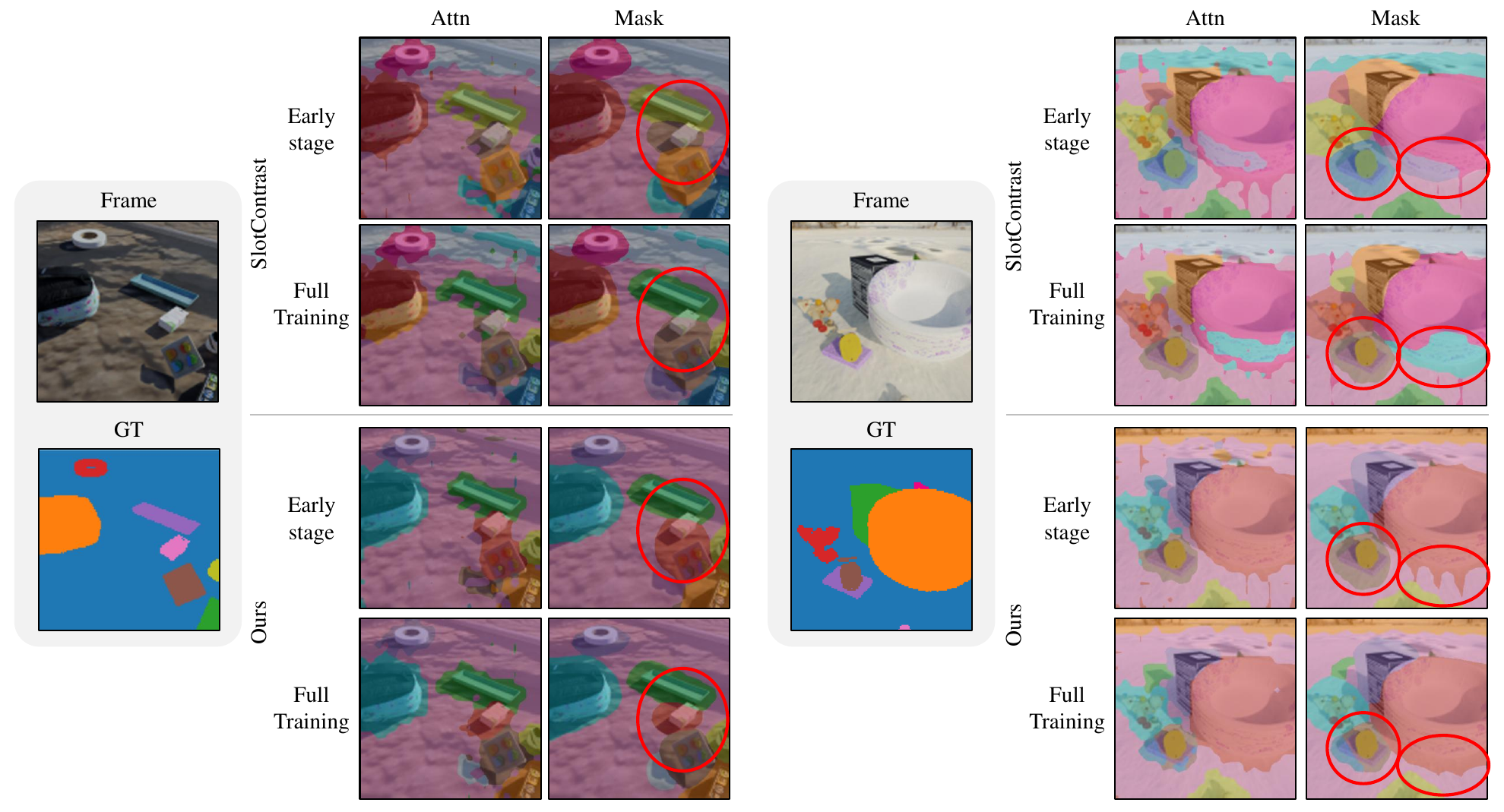} 
\caption{Qualitative analysis for vicious cycle.
}
\label{fig.supple.vicious_cycle}
\end{figure*}

To investigate whether the vicious cycle between attention noise and mask blur indeed arises during training, we conduct a qualitative analysis on the MOVi-C dataset by visualizing both attention maps and masks at the early and converged stages of training. The visualizations are provided in Fig.~\ref{fig.supple.vicious_cycle}.

For SlotContrast, we observe that the quality of both attention and masks can deteriorate as training progresses.
In the left example of Fig.~\ref{fig.supple.vicious_cycle}, objects that are initially well separated gradually lose their semantic boundaries, causing multiple objects to be merged into a single slot~(red circle).
In the right example, blurred boundaries prevent the model from disentangling overlapping objects, and residual attention noise persists even after training, propagating into the decoder masks.

In contrast, our method effectively suppresses this error propagation.
In the left example, even when some objects are under-segmented at early stages, the deblurring process of semantic boundaries encourages the model to recover clear object-wise separation as training proceeds.
In the right example, although the encoder attention maps initially exhibit noisy and blurred boundaries, our method progressively removes this noise and yields sharper encoder attention and cleaner decoder masks by the end of training.

% \textcolor{blue}{
% Specifically, as the model updates its masks using noisy attention responses, the masks gradually accumulate artifacts and inconsistencies, amplifying the noise in the attention maps themselves. 
% This mutually reinforcing effect becomes more pronounced over time, illustrating the vicious cycle. Moreover, due to ambiguous or blurred object boundaries in the input, the baseline struggles to delineate semantic regions, often producing under-fragmented masks in which two objects are merged into a single region.
% }

% \textcolor{blue}{
% In contrast, our method effectively prevents such error propagation. The predicted masks remain stable throughout the training, and the corresponding attention maps exhibit significantly reduced noise. Importantly, our approach demonstrates stronger discrimination at semantic boundaries: regions that were previously under-fragmented in the early training stage become correctly separated, enabling the model to assign distinct object slots. These results confirm that our method mitigates the vicious cycle and yields more reliable object-centric representations.
% }

\renewcommand{\thesection}{C}
\renewcommand{\thetable}{C\arabic{table}}
\renewcommand{\thefigure}{C\arabic{figure}}
\setcounter{table}{0}
\setcounter{figure}{0}

\section{Broader Impacts}
The advancements presented in this work have significant potential for positive societal impact by enhancing the capabilities of machines to understand and interact with the dynamic world in a more human-like, object-centric manner. 
By enabling robust unsupervised object discovery and tracking, our SRL can power more effective, efficient, and accessible tools for a wide range of applications without requiring costly human annotations.

However, the improved capabilities for unsupervised object tracking and segmentation could be repurposed for malicious uses.
A primary concern is the potential for enhanced surveillance and monitoring. 
A system that can reliably identify and track distinct objects without supervision could be deployed in mass surveillance systems without the subject's consent, raising significant privacy concerns.

In addition, the synergistic refinement, which is the core principle of our work, suggests a generalizable paradigm for other foundational architectures beyond object-centric learning, where the framework consists of encoder-decoder architectures.
For instance, the training dynamics of Generative Adversarial Networks~(GANs) exhibit a similar discrepancy between the representations of the discriminator and the generator.
While the generator's features are semantically coherent enough to produce segmentation masks~\citep{datasetgan}, the discriminator has been observed to lose the semantic information as training progresses~\citep{selfsupgan}.
Yet, the discriminator learning useful semantics has proven beneficial for stable GAN training~\citep{selfsupgan}.
Therefore, we posit that the feature discrepancy in encoder-decoder architectures can be leveraged as a complementary training signal in other domains as well.

%%%%%%%%%%%%%%%%%%%%%%%%%%%%%%%%%%%%%%%%%%%%%%%%%%%%%%%%%%%%%%%%%%
%%%%%%%%%%%%%%%%%%%%%%%%%%%%%%%%%%%%%%%%%%%%%%%%%%%%%%%%%%%%%%%%%%

\renewcommand{\thesection}{D}
\renewcommand{\thetable}{D\arabic{table}}
\renewcommand{\thefigure}{D\arabic{figure}}
\setcounter{table}{0}
\setcounter{figure}{0}

\section{The Use of Large Language Models~(LLMs)}
We used an LLM-based writing assistant solely for language refinement, including grammar correction, phrasing improvements, and ensuring clarity.
The model did not generate ideas, analyses, experiments, or results. 
All technical content was authored and verified by the authors, who take full responsibility for the manuscript. 
We affirm that no proprietary data beyond the text itself was shared with the writing tool.

\end{document}